\documentclass{article}

\PassOptionsToPackage{numbers, compress}{natbib}
\usepackage[final]{neurips_2022}

\usepackage[utf8]{inputenc} 
\usepackage[T1]{fontenc}    
\usepackage[pagebackref=true,breaklinks=true,colorlinks,bookmarks=false]{hyperref}
\usepackage{url}            
\usepackage{booktabs}     
\usepackage{amsfonts}      
\usepackage{nicefrac}    
\usepackage{microtype}  
\usepackage{color}
\usepackage{xcolor}        
\usepackage{enumerate}
\usepackage{tikz}
\usepackage{comment}
\usepackage{graphicx}
\usepackage{amsmath,amssymb} 
\usepackage{array}

\usepackage{subcaption}

\usepackage{tikz}
\usepackage{multirow}
\usepackage{algorithm}
\usepackage{algorithmic}
\DeclareMathOperator*{\argmax}{argmax}
\DeclareMathOperator*{\argmin}{argmin}

\usepackage[accsupp]{axessibility}  

\definecolor{NVblue}{rgb}{0.07,0.12,0.83}
\definecolor{BUred}{rgb}{0.8,0.0,0.0}

\newcommand{\std}[1]{\tiny{$\pm$ #1}}

\title{Efficient and Effective Augmentation \\Strategy for Adversarial Training}

\author{%
  Sravanti Addepalli $^\dagger$ \thanks{Equal Contribution. Correspondence to Sravanti Addepalli $<$sravantia@iisc.ac.in$>$, Samyak Jain $<$samyakjain.cse18@itbhu.ac.in$>$ $^\ddagger$ Work done during internship at Video Analytics Lab, Indian Institute of Science.}  \qquad Samyak Jain $^\dagger$ $^\diamond$ $^\ddagger$ \footnotemark[1] \qquad R.Venkatesh Babu $^\dagger$ \\
  $^\dagger$ Video Analytics Lab, Indian Institute of Science, Bangalore  \\ $^\diamond$ Indian Institute of Technology (BHU) Varanasi \hfill}

\begin{document}

\maketitle

\begin{abstract}
Adversarial training of Deep Neural Networks is known to be significantly more data-hungry when compared to standard training. Furthermore, complex data augmentations such as AutoAugment, which have led to substantial gains in standard training of image classifiers, have not been successful with Adversarial Training. We first explain this contrasting behavior by viewing augmentation during training as a problem of domain generalization, and further propose Diverse Augmentation-based Joint Adversarial Training (DAJAT) to use data augmentations effectively in adversarial training. We aim to handle the conflicting goals of enhancing the diversity of the training dataset and training with data that is close to the test distribution by using a combination of simple and complex augmentations with separate batch normalization layers during training. We further utilize the popular Jensen-Shannon divergence loss to encourage the \emph{joint} learning of the \emph{diverse augmentations}, thereby allowing simple augmentations to guide the learning of complex ones. Lastly, to improve the computational efficiency of the proposed method, we propose and utilize a two-step defense, Ascending Constraint Adversarial Training (ACAT), that uses an increasing epsilon schedule and weight-space smoothing to prevent gradient masking. The proposed method DAJAT achieves substantially better robustness-accuracy trade-off when compared to existing methods on the RobustBench Leaderboard on ResNet-18 and WideResNet-34-10. 
The code for implementing DAJAT is available here: \url{https://github.com/val-iisc/DAJAT}.

\end{abstract}

\vspace{-0.3cm}

\section{Introduction}

\label{sec:intro}

Deep Neural Network (DNN) based image classifiers are vulnerable to crafted imperceptible perturbations known as Adversarial Attacks \cite{intriguing-iclr-2014} that can flip the predictions of the model to unrelated classes leading to disastrous implications. Adversarial Training \cite{goodfellow2014explaining,madry-iclr-2018,zhang2019theoretically} has been the most successful defense strategy, where a model is explicitly trained to be robust in the presence of such attacks.
While early defenses focused on designing suitable loss functions for training, subsequent works \cite{pang2020bag,rice2020overfitting} showed that with careful hyperparameter tuning, even the two most popular methods PGD-AT \cite{madry-iclr-2018} and TRADES \cite{zhang2019theoretically} yield comparable performance, highlighting the saturation in performance with respect to changes in the training loss. Schmidt et al. \cite{schmidt2018adversarially} observed that adversarial training has a large sample complexity and further gains require the use of additional training data. Subsequent works \cite{carmon2019unlabeled,gowal2021improving} indeed used additional data whose distribution is close to that of the original dataset in order to obtain performance gains. The availability of large amounts of relevant data is impractical to assume, leading to an exploration towards augmentations based on Generative Adversarial Networks \cite{goodfellow2014generative} and Diffusion based models \cite{ho2020denoising,gowal2021improving}. However, the use of such generative models incurs an additional training cost and suffers from limited diversity in low-data regimes and in datasets with high-resolution images. 

A simple and efficient solution to improve the diversity of training data in standard Empirical Risk Minimization (ERM) based training has been the use of random transformations such as rotation, color jitter, and variations in contrast, sharpness and brightness \cite{krizhevsky2012imagenet,cubuk2018autoaugment,cubuk2020randaugment}, which can change images significantly in input space while belonging to the same class as the original image. However, prior works have surprisingly found that such augmentations, that cause large changes in the input distribution, do not help adversarial training \cite{rice2020overfitting,gowal2020uncovering,stutz2021relating}. This limits the augmentations in adversarial training to simple ones - zero padding followed by random crop, and horizontal flip \cite{rice2020overfitting,pang2020bag,gowal2020uncovering} - which may not be able to fill in the large data requirement of Adversarial Training.  

\begin{figure}
\centering
        \includegraphics[width=1\linewidth]{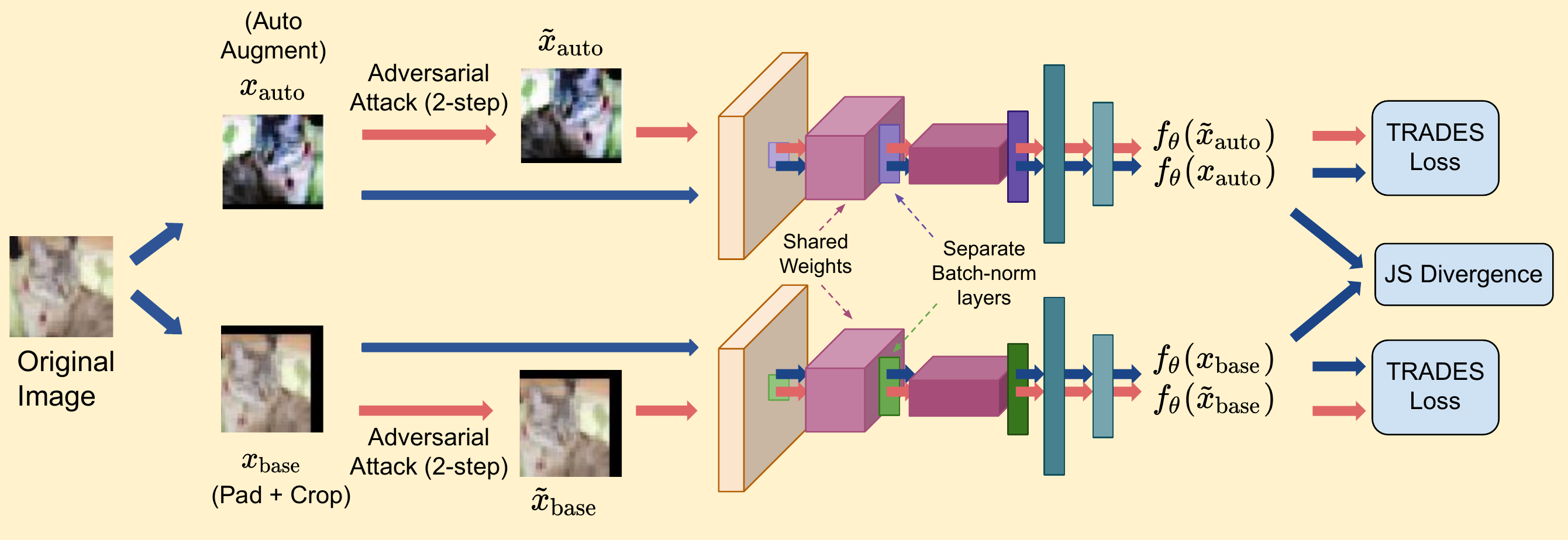}
        \caption{A Schematic representation of the proposed approach DAJAT}
        \label{fig:dajat_diag}
        \vspace{-0.3cm}
\end{figure}

In this work, we firstly analyze the reasons for this contrasting trend between Standard and Adversarial Training, and further show that \textbf{it is indeed possible to utilize complex augmentations effectively in Adversarial training as well}, by jointly training on simple and complex data augmentations using separate batch-normalization layers for each type, as shown in Fig.\ref{fig:dajat_diag}. While complex augmentations increase the data diversity resulting in better generalization, simple augmentations ensure that the model specializes on the training data distribution as well. We further minimize the Jenson-Shannon divergence between the softmax outputs of various augmentations to enable the simple augmentations to guide the learning of complex ones. In order to improve the computational efficiency of the proposed method, we use two attack steps (instead of 10) during training. By progressively increasing the magnitude of perturbations and performing smoothing in weight space, we show that it is indeed possible to improve the stability of training. Our contributions are listed below:
\begin{itemize}
\itemsep0em
\vspace{-0.1cm}
\item We analyze the reasons for the failure of strong data augmentations in adversarial training by viewing augmentation during training as a domain generalization problem, and further propose \textit{Diverse Augmentation based Joint Adversarial Training} (DAJAT) to utilize data augmentations effectively in Adversarial training. The proposed approach can be integrated with many augmentations and adversarial training methods to obtain performance gains. 
\item We propose and integrate DAJAT with an efficient 2-step defense strategy, \textit{Ascending Constraint Adversarial Training} (ACAT) that uses linearly increasing $\varepsilon$ schedule, cosine learning rate and weight-space smoothing to prevent gradient masking and improve convergence. 
\item We obtain improved robustness and large gains in standard accuracy on multiple datasets (CIFAR-10, CIFAR-100, ImageNette) and model architectures (RN-18, WRN-34-10). We obtain remarkable gains in a low data scenario (CIFAR-100, Imagenette) where data augmentations are most effective. \textbf{On CIFAR-100, we outperform all existing methods on the RobustBench leaderboard \cite{croce2021robustbench} with the same model architecture.}

\end{itemize}

\vspace{-0.3cm}
\section{Related Works}
We discuss various existing strategies for improving the Adversarial robustness of Deep Networks. 
\noindent \textbf{Adversarial Training (AT):} Goodfellow et al. \cite{goodfellow2014explaining} proposed FGSM-AT where single-step adversarial samples were used for training. However, these models were susceptible to gradient masking \cite{papernot2017practical}, where the local loss landscape becomes convoluted leading to the generation of weaker single-step attacks during training. This leads to a false sense of security against single-step attacks, while the models are still susceptible to stronger multi-step attacks such as PGD \cite{madry-iclr-2018}. PGD-AT \cite{madry-iclr-2018, rice2020overfitting} used multi-step attacks in a similar adversarial training formulation to obtain robust models that stood the test of time against several attacks \cite{athalye2018obfuscated,croce2020reliable,sriramanan2020gama}. TRADES \cite{zhang2019theoretically} explicitly optimizes the trade-off between the accuracy on natural and adversarial examples by minimizing the cross-entropy loss on natural images along with the Kullback-Leibler (KL) divergence between the predictions of adversarial and clean images. In the proposed defense, we use the base loss from TRADES-AT \cite{zhang2019theoretically}.

Several works have explored the use of auxiliary techniques in AT such as weight-space smoothing \cite{izmailov2018averaging, wu2020adversarial}, architectural changes \cite{gowal2020uncovering,xie2020smooth} and increasing the diversity of training data by using additional natural and synthetic data \cite{schmidt2018adversarially,carmon2019unlabeled,gowal2020uncovering,rebuffi2021data}. Increasing the diversity of training data achieves significant performance gains since the sample complexity of adversarial training is known to be high \cite{schmidt2018adversarially}. 

\noindent \textbf{Augmentations in Adversarial Training:} 
While data augmentations such as contrast, sharpness and brightness adjustments are known to improve performance in the standard training regime, they have not led to substantial gains in adversarial training. AutoAugment \cite{cubuk2018autoaugment} uses Proximal Policy Optimization to find the set of policies that can yield optimal performance on a given dataset. Contrary to prior works \cite{rice2020overfitting,gowal2020uncovering}, we show that the policies that are optimized for standard training indeed yield a boost in performance of adversarial training as well, when used in the proposed training framework. 

In a recent work, Rebuffi et al. \cite{rebuffi2021data} show that it is possible to obtain substantial gains in robust accuracy by using spatial composition based augmentations such as CutMix \cite{yun2019cutmix} and CutOut \cite{cutout} that preserve low-level features of the image. Cutmix replaces part of an image with another and also combines the output softmax vectors in the same ratio, while CutOut blanks out a random area of an image. The authors hypothesize that the augmentations used in Adversarial training need to preserve low-level features, which severely limits the possibilities for mitigating the large sample complexity in adversarial training. We show that by using the proposed approach DAJAT, it is indeed possible to use augmentations such as color jitter, contrast, sharpness and brightness adjustments that significantly change the low level statistics of images (Ref: Appendix-\ref{app:simplecomplexaugs}).  
\vspace{-0.1cm}
\section{Preliminaries: Notation and Threat Model}
\vspace{-0.2cm}
We consider the Adversarial Robustness of DNN based image classifiers. An input image is denoted as $x \in \mathcal{X}$ and the corresponding ground truth label as $y \in [0,1]$. We denote a simple transformation of $x$ obtained using  Pad, Crop and Horizontal flip (Pad+Crop+HFlip, referred to as Base augmentations) by $x_{\mathrm{base}}$ and other transformations of $x$ by the respective subscript. For example, $x_{\mathrm{auto}}$ refers to the image $x$ being transformed by AutoAugment (AA) \cite{cubuk2018autoaugment} followed by the base augmentations. The function mapping of the classifier $C$ from input space $\mathcal{X}$ to the softmax vectors is denoted using $f_{\theta}(.)$, where $\theta$ denotes the network parameters. Adversarial examples corresponding to the images $x$, $x_{\mathrm{base}}$ and $x_{\mathrm{auto}}$ are denoted using $\Tilde{x}$, $\Tilde{x}_{\mathrm{base}}$ and $\Tilde{x}_{\mathrm{auto}}$ respectively. We consider the $\ell_{\infty}$ norm based threat model, where $\tilde{x}$ is a valid perturbation within $\varepsilon$ if it belongs to the set $\mathcal{A}_\varepsilon(x) = \{\tilde{x}:||\tilde{x}-x||_\infty \leq \varepsilon\}$. 

\vspace{-0.1cm}
\section{Motivation: Role of Augmentations in Neural Network Training}
\label{sec:motivation}
\vspace{-0.2cm}

In this section, we first explore the contrasting factors that influence the training of neural networks when data is augmented (Sec.\ref{subsec:aug_NN}), and further delve into the specifics of adversarial training which make it challenging to obtain gains using complex data augmentations (Sec.\ref{subsec:aug_AT}). 

\vspace{-0.2cm}
\subsection{Impact of Augmentations in Neural Network Training}
\label{subsec:aug_NN}
\vspace{-0.1cm}
\textbf{Conjecture-1:} We hypothesize that the role of data augmentations in the training of Neural Networks is influenced by the following contrasting factors:
\vspace{-0.2cm}
\begin{enumerate}[(i)]
\item Reduced overfitting due to an increase in diversity of the augmented dataset, leading to better generalization of the network to the test set.
\item Larger domain shift between the augmented data distribution and the test data distribution, leading to a drop in performance on the test set.
\item Capacity of the Neural Network in being able to generalize well to the augmented data distribution and the unaugmented data distribution for the given task. 
\end{enumerate}

\vspace{-0.1cm}
\noindent \textbf{Justification:} The training of Neural Networks using augmented data can be considered as a problem of domain generalization, where the network is trained on a source domain (augmented data) and is expected to generalize to a target domain (test data). We use the theoretical formulation by  Ben-David et al. \cite{ben2010theory} shown below to justify the respective claims in Conjecture-1:
\addtolength{\abovedisplayshortskip}{-3ex}
\addtolength{\belowdisplayshortskip}{-5ex}
\begin{equation}
\label{bendavid}
\epsilon_t(f) \leq \epsilon_s(f) + \frac{1}{2}d_{\mathcal{F}\Delta\mathcal{F}}(s,t) + \lambda
\end{equation}
\begin{enumerate}[(i)]
\item The use of a more diverse or larger source dataset reduces overfitting, improving the performance of the network on the source distribution. From Eq.\ref{bendavid}, expected error on the target distribution $\epsilon_t$ (test set in this case) is upper bounded by the expected error on the source distribution $\epsilon_s$ (augmented dataset) along with other terms. \textcolor{NVblue}{Therefore, improved performance on the augmented distribution can improve the performance on the test set as well.} 
\item The expected error on the target distribution $\epsilon_t$ is upper bounded by the distribution shift between the source and  target distributions $\frac{1}{2}d_{\mathcal{F}\Delta\mathcal{F}}(s,t)$  along with other terms. \textcolor{NVblue}{Therefore, a larger domain shift between the augmented and test data distributions can indeed limit the performance gains on the test set.} 
\item The constant $\lambda$ in Eq.\ref{bendavid} measures the risk of the optimal joint classifier: $\lambda =\min\limits_{f\in\mathcal{F}} \epsilon_s(f) + \epsilon_t(f)$. Neural Networks with a higher capacity can minimize the expected risk on the source set $\epsilon_s$ and the risk of the optimal joint classifier effectively. \textcolor{NVblue}{Therefore, capacity of the Neural Network and complexity of the task influence the gains that can be obtained using augmentations.}
\end{enumerate}

\vspace{-0.3cm}
\subsection{Analysing the role of Augmentations in Adversarial Training}
\label{subsec:aug_AT}
\vspace{-0.1cm}
We analyse the trade-off between the factors described in Conjecture-1 for adversarial training when compared to standard ERM training. In addition to the goal of improving accuracy on clean samples, adversarial training aims to achieve local smoothness of the loss landscape as well. Hence, the complexity of Adversarial Training is higher than that of standard ERM training, making it important to use larger model capacities to obtain gains using data augmentations (based on Conjecture-1 (iii)). This justifies the gains obtained by Rebuffi et al. \cite{rebuffi2021data} on the WRN-70-16 architecture by using CutMix based augmentations  (2.9\% higher robust accuracy and 1.23\% higher clean accuracy). The same method does not obtain significant gains on smaller architectures such as ResNet-18 where a 1.76\% boost in robust accuracy is accompanied by a 2.55\% drop in clean accuracy. 

\begin{table*}
\begin{minipage}{0.46\linewidth}
\caption{\textbf{Impact of augmentations:} Performance (\%) of ACAT models on Base augmentations and AutoAugment (Auto). Clean and robust accuracy against GAMA attack \cite{sriramanan2020gama} are reported. The use of AutoAugment results in $\sim1.5-2\%$ drop in robust accuracy.}
\setlength\tabcolsep{2pt}
\resizebox{1.0\linewidth}{!}{
\label{table:baseauto}
\begin{tabular}{l|c|c|c|c|c}
\toprule
\textbf{}                          & \textbf{Test:}                         & \multicolumn{2}{c|}{\textbf{No Aug}}                                                     & \multicolumn{2}{c}{\textbf{AutoAugment}}                     \\
\cmidrule{2-6}
\textbf{Model}      & \textbf{Train $\downarrow$} & \multicolumn{1}{c|}{\textbf{Clean}} & \multicolumn{1}{c|}{\textbf{Robust}} & \multicolumn{1}{c|}{\textbf{Clean}} & \multicolumn{1}{c}{\textbf{Robust}} \\
\midrule

                                   & Base                                   & 82.41                                             & \textbf{50.00}                                              & 63.79                                             & 37.07                      \\
\multirow{-2}{*}{ResNet-18}        & Auto                                   & \textbf{82.54}                                             & 48.11                                              & \textbf{76.40}                                             & \textbf{43.22}                      \\
\midrule
                                   & Base                                   & 86.71                                             & \textbf{55.58}                                              & 68.24                                             & 40.83                      \\
                                
\multirow{-2}{*}{WideResNet-34-10} & Auto                                   & \textbf{86.80}                                             & 53.99                                              & \textbf{82.64}                                             & \textbf{48.98} \\

\bottomrule
\end{tabular}
}
 \vspace{-0.5cm}
\end{minipage}
\hfill
\begin{minipage}{0.52\linewidth}
\centering
\includegraphics[width=\linewidth]{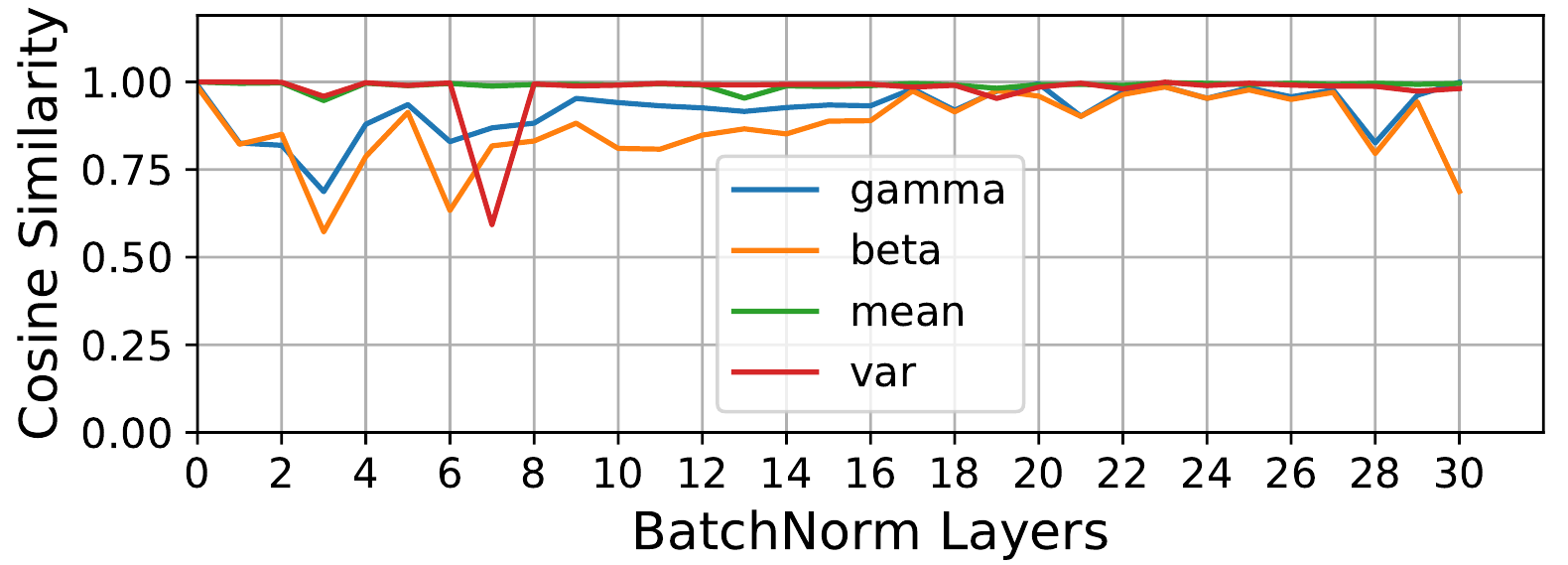}
 \vspace{-0.5cm}
\captionof{figure}{\textbf{Comparison of BN layer statistics} for a WRN-34-10 model trained on CIFAR-10 using DAJAT. BN layers of the Base augmentations (Pad+Crop,H-Flip) are compared with those of AutoAugment. Initial layer (L3) parameters are diverse, while those of deeper layers (L25) are similar.}
 \vspace{-0.5cm}
\label{fig:bn_plot}

\end{minipage}
        
\end{table*}

Secondly, while the distribution shift between augmented data and test data ($\frac{1}{2}d_{\mathcal{F}\Delta\mathcal{F}}(s,t)$) may be sufficiently low for natural images leading to improved generalization to test set (Conjecture-1(i,ii)), the same may not be true for adversarial images. There is a large difference between the augmented data and test data in pixel space, although they may be similar in feature space. Since adversarial attacks perturb images in pixel space, the distribution shift between the corresponding perturbations widens further as shown in Fig.\ref{fig:noaug}, \ref{fig:baseaug} and \ref{fig:autoaug}. Based on Conjecture-1(ii), unless this difference is accounted for, complex augmentations cannot improve the performance of adversarial training. (Ref. Appendix-\ref{app:simplecomplexaugs}). This trend has also been observed empirically by Rebuffi et al. \cite{rebuffi2021data}, based on which they conclude that the augmentations designed for robustness need to preserve low-level features. 

We present the performance of Adversarial Training by using either Base augmentations (Pad+Crop, Flip) or AutoAugment \cite{cubuk2018autoaugment} during training and inference on the CIFAR-10 dataset using ResNet-18 and WideResNet-34-10 architectures in Table-\ref{table:baseauto}. Firstly, we note that by using AutoAugment during training alone, robust accuracy on the test set drops by $\sim1.5-2\%$ which is as observed in prior work \cite{gowal2020uncovering}. Secondly, the clean and robust accuracy drop by $\sim6.5\%$ when augmented images are used for both training and testing, highlighting the complexity of the learning task. We present additional results by training without any augmentations, and by training using a combination of both augmentations in every minibatch in Table-\ref{table:acat_additional}. We note that the use of Base Augmentations alone (Pad+Crop+HFlip) still gives the best overall performance on the unaugmented test set. 

\vspace{-0.2cm}
 
\section{Proposed Method}

\vspace{-0.2cm}
\subsection{Background}
\label{subsec:trades}
We briefly discuss the TRADES-AWP defense \cite{zhang2019theoretically,wu2020adversarial}, which is the base algorithm used in DAJAT. 
\addtolength{\abovedisplayshortskip}{0ex}
\addtolength{\belowdisplayshortskip}{0ex}
\abovedisplayskip=-1pt
\belowdisplayskip=-1pt
\begin{equation}
\label{eq:awp}
    \mathcal{L}_{\mathrm{AWP}} = \max\limits_{\hat{\theta} \in \mathcal{M}(\theta)} \frac{1}{N} \sum\limits_{i=1}^N \mathcal{L}_\mathrm{CE}(f_{\theta + \hat{\theta}}(x_i),y_i) + \beta \cdot \max\limits_{\tilde{x}_i \in \mathcal{A}_\varepsilon(x_i)} \mathrm{KL}(f_{\theta + \hat{\theta}}(x_i)||f_{\theta + \hat{\theta}}(\tilde{x}_i))
\end{equation}

Firstly, an adversarial attack is generated by maximizing the KL divergence between the softmax predictions of the clean and adversarial examples iteratively for 10 attack steps. An Adversarial Weight Perturbation (AWP) step additionally perturbs the weights of the model to maximize the overall loss in weight-space. The weight perturbations are constrained in the feasible region $\mathcal{M}(\theta)$ such that for a given layer $l$, $||\hat{\theta_l}|| \leq \gamma \cdot ||\theta_l||$. The overall training loss is a combination of Cross-entropy loss on clean samples and the KL divergence term. 
The latter is weighted by a factor $\beta$ that controls the robustness-accuracy trade-off. Training the model using Adversarial Weight Perturbations leads to smoothing of loss surface in the weight space, resulting in better generalization \cite{wu2020adversarial,stutz2021relating}.

\vspace{-0.2cm}
\subsection{Diverse Augmentation based Joint Adversarial Training (DAJAT)}

As discussed in Section-\ref{sec:motivation}, the use of augmentations in training can be viewed as a problem of domain generalization, where performance on the source distribution or augmented dataset is crucial towards improving the performance on the target distribution or test set. Since adversarial training is inherently challenging, for limited model capacity, it is difficult to obtain good performance on the training data that is transformed using complex augmentations. Moreover, the large distribution shift between augmented data and test data, specifically with respect to low-level statistics, results in poor generalization of robust accuracy to the test set. 

To mitigate these challenges, we propose the combined use of simple and complex augmentations during training, so that the model can benefit from the diversity introduced by complex augmentations, while also specializing on the original data distribution that is similar to the simple augmentations. We propose to use separate batch normalization layers for simple and complex augmentations, so as to offset the shift in distribution between the two kinds of augmentations. In our main approach, we propose to use Pad and Crop followed by Horizontal Flip (Pad+Crop+HFlip) as the Simple augmentations, and Autoaugment followed by Pad+Crop+HFlip as the complex augmentations. We justify the choice of this augmentation pipeline in Appendix-\ref{appsubsec:justifyaugs}. 

Motivated by AugMix \cite{hendrycks2020augmix}, we additionally minimize the Jenson-Shannon (JS) divergence between the softmax outputs of different augmentations, so as to allow the simple augmentations to guide the learning of complex ones. We present the training loss $\mathcal{L}_{\mathrm{DAJAT}}$ of the proposed method, Diverse Augmentation based Joint Adversarial Training (DAJAT) below in Eq.\ref{eq:dajat_loss_eq}:

\abovedisplayskip=-7pt
\begin{equation}
\label{eq:trades}
    \mathcal{L}_{\mathrm{TR}}(\theta,x,y) =  \mathcal{L}_\mathrm{CE}(f_{\theta}(x),y) + \beta \cdot \max\limits_{\tilde{x} \in \mathcal{A}_\varepsilon(x)} \mathrm{KL}(f_{\theta}(x)||f_{\theta}(\tilde{x}))~~~~~~~~~~~~~~~~~~~~~~~~~~~~~~~~~~~~~~~~~~~~~
\end{equation}

\abovedisplayskip=-5pt
\belowdisplayskip=-1pt
\begin{equation}
\label{eq:awp_step}
   \tilde{\theta} = \argmax\limits_{\hat{\theta} \in \mathcal{M}(\theta)} \frac{1}{N} \sum\limits_{i=1}^N \mathcal{L}_{\mathrm{TR}}(\theta+\hat{\theta},x_{i,\mathrm{base}},y_i)~~~~~~~~~~~~~~~~~~~~~~~~~~~~~~~~~~~~~~~~~~~~~~~~~~~~~~~~~~~~~~~~~~~~~~~~~~~~~~~~~~~
\end{equation}

\abovedisplayskip=-5pt
\belowdisplayskip=1pt
\begin{multline}
\label{eq:dajat_loss_eq}
  \mathcal{L}_{\mathrm{DAJAT}} =  \frac{1}{T+1} \cdot \frac{1}{N} \sum\limits_{i=1}^N \big\{ \mathcal{L}_{\mathrm{TR}}(\theta+\tilde{\theta},x_{i,\mathrm{base}},y_i) +  \sum\limits_{t=1}^T  \mathcal{L}_{\mathrm{TR}}(\theta+\tilde{\theta},x_{i,\mathrm{auto(t)}},y_i)\big\} + \\  \frac{1}{N} \sum\limits_{i=1}^N \mathrm{JSD}(f_{\theta+\tilde{\theta}}(x_{i,\mathrm{base}}),f_{\theta+\tilde{\theta}}(x_{i,\mathrm{auto(1)}}), \dots , f_{\theta+\tilde{\theta}}(x_{i,\mathrm{auto(T)}}))
\end{multline}

Adversarial attacks are generated individually for each augmentation by maximizing the respective KL divergence term of the TRADES loss shown in Eq.\ref{eq:trades}. To improve training efficiency, we compute $\tilde{x}$ using two attack steps with a step-size of $\varepsilon$. We use a combination of a linearly increasing schedule of $\varepsilon$, cosine learning rate schedule and model weight-averaging \cite{izmailov2018averaging} to improve the stability and performance of adversarial training (Details in Sec.\ref{sec:ACAT}). The DAJAT loss (Eq.\ref{eq:dajat_loss_eq}) is a combination of the TRADES 2-step loss on each of the augmentations $x_\mathrm{base}$ and $x_\mathrm{auto(t)}$, along with an adversarial weight perturbation step on the loss corresponding to the base augmentations alone to reduce computational cost. For every batch normalization layer, two sets of running statistics and affine parameters are maintained and used for simple and complex augmentations respectively (Ref: Algorithm-\ref{alg:DAJAT}).

While we use AutoAugment in our main approach, we show in Table-\ref{table:aug} that the proposed approach works well with other augmentations as well. The role of the base (primary) augmentations is primarily to learn the batch normalization layers that would be used during inference time, and to provide better supervision for the training of complex augmentations using the JS divergence term. The role of the complex (secondary) augmentations is to enhance the diversity of the training dataset. Therefore we use a single primary augmentation and multiple instances of the secondary augmentation. We find that the gains in performance saturate with the addition of more instances of secondary augmentation, and therefore recommend the use of a single base augmentation and two instances of secondary augmentation for the best performance-accuracy trade-off. We note from Table-\ref{table:sota_c10} and Appendix-\ref{app:flops} that in this setting, the computational efficiency of the proposed method is better than the TRADES-AWP \cite{zhang2019theoretically,wu2020adversarial} defense, while achieving considerable performance gains. 

\subsection{Split Batch Normalization Layers for Different Augmentations}
\vspace{-0.1cm}

A Batch Normalization (BN) layer \cite{pmlr-v37-ioffe15} is implemented as follows on a given feature map $g(x_i)$ of an input image $x_i$: $\hat{g}(x_i) = \frac{g(x_i) - \mu}{\sigma} \cdot \gamma + \beta$. 
Here, $\mu$ and $\sigma$ denote the mean and standard deviation of the current mini-batch during training. During inference, these are set to the running mean and variance computed during training. $\gamma$ and $\beta$ constitute parameters of the network that are trained. 

Prior works use separate batch normalization layers to improve the performance of standard \cite{Xie_2020_CVPR,wangAugmax,Merchant2020DoesDA} and adversarial training \cite{Xieintriguing, wang2020onceforall}, in both supervised \cite{Xie_2020_CVPR,wangAugmax,Merchant2020DoesDA,Xieintriguing,wang2020onceforall} and self-supervised settings \cite{acl,advcl}. 
The proposed defense DAJAT uses separate batch normalization layers for simple and complex augmentations as discussed in the previous section. We do not use separate batch normalization for clean and adversarial images here.  

In DAJAT, we maintain two sets of BN statistics $\mu$ and $\sigma$, and two sets of affine parameters, $\beta$ and $\gamma$ for every BN layer. We plot the cosine similarity between the BN layers corresponding to Base augmentations and AutoAugment across different layers on a WideResNet-34-10 model trained using DAJAT on CIFAR-10 in Fig.\ref{fig:bn_plot}. While the mean and variance are relatively more similar across most of the layers (0-30 in x-axis), we note significant differences in the $\gamma$ and $\beta$ values, specifically in the initial layers. We present a detailed study on the importance of having separate running statistics and affine parameters in Appendix-\ref{app:splitbn}, and conclude that our method works effectively by separating either the running statistics or affine parameters alone as well. However, it works best by using a combination of both (Table-\ref{table:bn_additonal_one}). We further note that very small differences in running statistics (as seen in Fig.\ref{fig:bn_plot} and \ref{fig:bn_noaffine}) can also result in noteworthy performance gains. Finally, we find that the use of a single batch-norm layer for both augmentations degrades results significantly. 
Therefore, the use of different parameters for each augmentation type ensures that the function mapping differs for each augmentation, thereby effectively offsetting the large differences in low level statistics. 
\subsection{Ascending Constraint Adversarial Training (ACAT)}
\label{sec:ACAT}

\vspace{-0.1cm}
In this section, we discuss the methods incorporated to improve the training efficiency of DAJAT in greater detail. We apply these methods to the TRADES-AWP defense to independently analyse their impact, and term the proposed defense as Ascending Constraint Adversarial Training (ACAT). We aim to improve the training efficiency by reducing the number of attack steps of the base defense from 10 to 2. We use two attack steps for training since it is known to be more stable when compared to single-step adversarial training, while still being computationally efficient \cite{sriramanan2021nuat}. 

\begin{figure}
\centering
\includegraphics[width=1.0\linewidth]{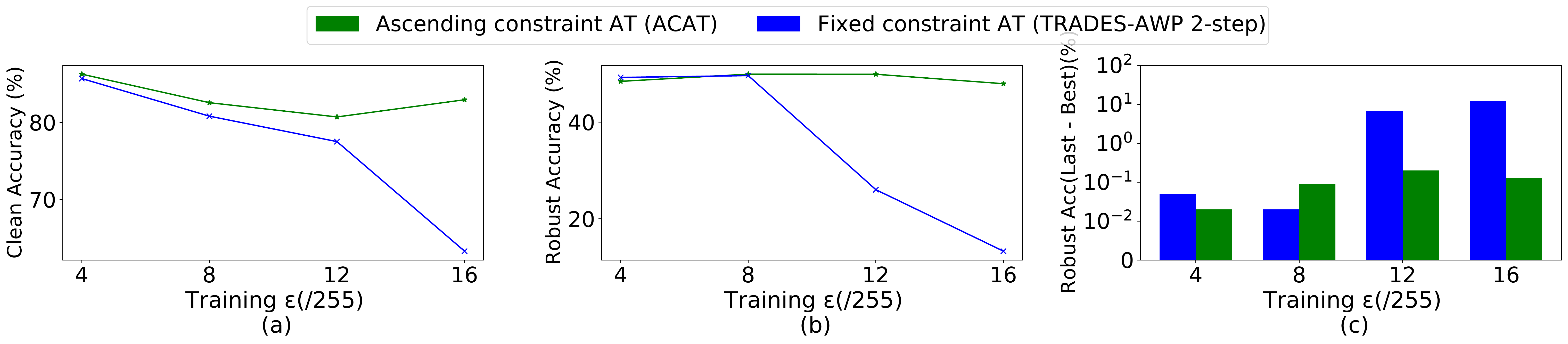}
\caption{\textbf{Comparison of ACAT (2-step) against TRADES-AWP 2-step baseline} on CIFAR-10 with ResNet-18 architecture. (a) Clean accuracy, (b) Robust accuracy, and (c) Difference in Robust accuracy between the last and best epochs are reported. ACAT has better performance and stability even at large training $\varepsilon$ values. Robust Accuracy is reported against GAMA attack \cite{sriramanan2020gama} on $\varepsilon=8/255$.}
\vspace{-0.8cm}
\label{fig:vareps}
\end{figure}

As shown in Fig.\ref{fig:vareps}, naively reducing the number of attack steps to 2 in TRADES-AWP AT (Fixed constraint AT) causes a drop in clean and robust accuracy. While the drop is larger at higher training $\varepsilon$, a drop in clean accuracy is seen at $\varepsilon=8/255$ as well. Further, the large robustness gap between last and best epochs indicates that the training stability deteriorates towards the end of training. The instability of TRADES-AWP AT at $\varepsilon=8/255$ is higher for larger model capacities (WideResNet-34-10) as discussed in Appendix-\ref{appsubsec:motivacat}.

Curriculum learning approaches \cite{Sitawarin2020ImprovingAR, CaiCAT} have been used to improve robustness-accuracy trade-off in adversarial training. Prior works \cite{shaeiri2020towards,addepalli2021oaat} also show that training convergence at large $\varepsilon$ bounds can be improved by linearly increasing $\varepsilon$ as training progresses. Inspired by this, we propose to linearly increase $\varepsilon$ alongside a cosine learning rate schedule and weight-space smoothing for improving the stability and convergence of training (Ref: Algorithm-\ref{alg:ACAT}). The use of a lower learning rate at larger $\varepsilon$ improves training stability. As shown in Fig.\ref{fig:vareps} and Table-\ref{table:vareps_abl_two}, the performance and stability of the ACAT are significantly better when compared to the TRADES-AWP 2-step baseline, specifically at larger perturbation bounds and higher model capacities, at the same computational cost. 
The proposed defense maintains a good clean accuracy at all the training $\varepsilon$ values considered, and has $<0.1\%$ robustness difference between last and best epochs.

We thus show that even without modifying the training loss function it is possible to obtain significant gains in the stability and performance of 2-step adversarial training by using the proposed Ascending Constraint Adversarial Training. The proposed method can indeed improve the performance of other defenses such as GAT \cite{sriramanan2020gama} and NuAT \cite{sriramanan2021nuat} as well, as shown in Table-\ref{table:acat_gat_nuat}. 

\vspace{0.2cm}

\noindent \textbf{Analysis of the proposed ACAT defense:} 

\noindent \textbf{Conjecture-2:} We hypothesize that given a Neural Network $f_{\theta_\varepsilon}$ that minimizes the TRADES loss \cite{zhang2019theoretically} within a maximum perturbation radius of $\varepsilon$, and has sufficient smoothness in weight space within an $\ell_2$ radius $\psi$ around $\theta_\varepsilon$, there exist $\varepsilon’ > \varepsilon$ and a model $f_{\theta_{\varepsilon'}}$ where $||\theta_{\varepsilon'} - \theta_\varepsilon|| \leq \psi$, such that $f_{\theta_{\varepsilon'}}$  has a lower TRADES loss within $\varepsilon'$ when compared to $f_{\theta_{\varepsilon}}$. 

\textbf{Justification} presented in Section-\ref{subsec:acat_analysis_supple}.

Based on Conjecture-2, by selecting a small enough learning rate $\eta$ and large enough $\psi$, it is possible to move towards a better solution w.r.t. the TRADES loss at $\varepsilon'$. The smoothness of loss surface further ensures that the gradients at $\varepsilon'$ are bounded with a low magnitude, preventing the onset of catastrophic overfitting, which is known to be associated with large magnitude gradients \cite{li2021subspace}. This motivates the proposed approach ACAT that effectively combines an increasing $\varepsilon$ schedule, cosine learning rate schedule that allows low learning rates at larger $\varepsilon$ values, and adversarial weight perturbation to ensure smoothness in the loss surface. 

\vspace{0.1cm}
\noindent \textbf{Stability improvements obtained using ACAT:} We consider an extreme case of using single-step attacks for training, to demonstrate the stability improvements obtained using ACAT. We consider a checkpoint obtained at $\varepsilon=7.25/255$ while training using the proposed ACAT defense on CIFAR-10 dataset for a WideResNet-34-10 architecture. We use this model as an initialization and perform single-step adversarial training using the TRADES-AWP training objective at $\varepsilon$=$7.5/255$. 
Our goal here is to merely verify the stability of single-step training, although the use of single-step attack makes the robustness gains marginal. We present the clean accuracy and robust accuracy against PGD-20 attack using different values of Adversarial Weight Perturbation step size $\gamma$ in Fig.\ref{fig:gamma}(a). It is evident from the plots that for a fixed value of learning rate, increasing $\gamma$ can delay the onset of gradient masking, thereby improving the training stability. We also note marginal improvements in performance at higher $\gamma$, due to the enhanced generalization. Therefore, weight-space smoothing indeed helps in improving the performance and stability of the proposed defense. We also note from Fig.\ref{fig:gamma}(b) that for a fixed value of $\gamma=0.01$, lower learning rate leads to better stability, therefore cosine schedule also helps in improving stability.
We note that the value of gamma in relation to the learning rate is crucial. Therefore, in the proposed approach, we use a fixed value of $\gamma$ and reduce the learning rate using a cosine learning rate schedule to improve training stability as $\varepsilon$ increases. 
\vspace{-0.2cm}
\section{Experiments and Results}
\vspace{-0.2cm}
\subsection{State-of-the-art comparison}
\vspace{-0.2cm}
We compare the performance of the proposed defenses ACAT and DAJAT against the 2-step and 10-step state-of-the-art defenses NuAT2-WA \cite{sriramanan2020gama}, TRADES-AWP \cite{zhang2019theoretically,wu2020adversarial} and PGD-AT \cite{madry-iclr-2018,rice2020overfitting} in Tables- \ref{table:sota_c10}, \ref{table:sota_c100} and \ref{table:200_epochs} on the CIFAR-10, CIFAR-100 \cite{krizhevsky2009learning} and ImageNette \cite{howard2020fastai,imagenet_cvpr09} datasets. Although there have been several defenses after this, we show in Table-\ref{table:contemporary_works} that TRADES-AWP outperforms them in most settings, and the proposed defense DAJAT consistently outperforms all methods, showing substantial gains in performance. Further, since we integrate the proposed approach with TRADES-AWP, we primarily compare our approach with this across all datasets. We show in Table-\ref{table:dajat_with_others} that DAJAT can be integrated with other base defenses \cite{addepalli2021oaat,rade2022reducing} as well to obtain performance gains. We integrate model weight averaging with the TRADES-AWP defense (termed as TRADES-AWP-WA) as well for a fair comparison. We train all models for 110 epochs unless specified otherwise.
\begin{table}
\caption{\textbf{CIFAR-10:} Performance (\%) of the proposed defenses ACAT and DAJAT when compared to the state-of-the-art. Robust evaluations are performed on GAMA \cite{sriramanan2020gama} and AutoAttack \cite{croce2020reliable}. Training time per epoch is reported by running each algorithm across 2 V100 GPUs.}
\setlength\tabcolsep{2pt}
\resizebox{1.0\linewidth}{!}{
\label{table:sota_c10}
\begin{tabular}{l|c|c|c|c|c||c|c|c}
\toprule
 \multicolumn{1}{l|}{} & \multicolumn{1}{c|}{} & \multicolumn{4}{c||}{\textbf{CIFAR-10, ResNet-18}}                                                                                                                                                    & \multicolumn{3}{c}{\textbf{CIFAR-10, WideResNet-34}}                                                                                                             \\
 \cmidrule{3-9}
\multicolumn{1}{l|}{\textbf{Method}} & \multicolumn{1}{c|}{\textbf{~~~Steps~~~}} & \multicolumn{1}{c|}{\textbf{~~Clean~~}} & \multicolumn{1}{c|}{\textbf{~~GAMA~~}} & \multicolumn{1}{c|}{\textbf{AutoAttack}} & \textbf{Time/epoch (sec)} & \multicolumn{1}{c|}{\textbf{~~Clean~~}} & \multicolumn{1}{c|}{\textbf{~~GAMA~~}} & \multicolumn{1}{c}{\textbf{AutoAttack}} \\
    \midrule
NuAT2-WA  \cite{sriramanan2021nuat}    & 2                                                       & 82.21                                                      & \textbf{50.97}                                                     & \textbf{50.75}                                                           & 109                              & 86.32                                                      & 55.08                                                     & 54.76                                                           \\
ACAT, Ours (Base, 2step)  & 2                                                 & \textbf{82.41}                                                      & 50.00                                                     & 49.80                                                           & 95                               & \textbf{86.71}                                                      & \textbf{55.58}                                                     & \textbf{55.36}                                                           \\
\midrule
PGD-AT \cite{madry-iclr-2018}            & 10                                                  & 81.12                                                      & 49.08                                                     & 48.75                                                           & 182                              & 86.07                                                      & 52.70                                                     & 52.19                                                           \\
TRADES-AWP \cite{zhang2019theoretically,wu2020adversarial}          & 10                                                & 80.47                                                      & 50.06                                                     & 49.87                                                           & 228                              & 85.19                                                      & 55.87                                                     & 55.69                                                           \\
TRADES-AWP-WA                & 10                                        & 80.41                                                      & 49.89                                                     & 49.67                                                           & 228                              & 85.10                                                      & 56.07                                                     & 55.87                                                           \\
TRADES-AWP-WA (200 epochs)~~~~~~~~~~ & 10 & 81.99                                                      & 51.65                                                     & 51.45                                                           & 228                              & 85.36                                                      & 56.35                                                     & 56.17                                                           \\
DAJAT, Ours (Base, AA)     & 2 + 2                                           & \multicolumn{1}{c|}{85.60}          & \multicolumn{1}{c|}{51.27}         & 51.06                                                           & 160                              & 87.87                                                      & 56.97                                                     & 56.68                                                           \\
DAJAT, Ours (Base, 2*AA)               & 2  + 4                            & 85.99                                                      & 51.71                                                     & 51.48                                                           & 219                              & \textbf{88.90}                                                      & 57.22                                                     & 56.96                                                           \\
DAJAT, Ours (Base, 3*AA)                  & 2 + 6                            & \textbf{86.67}                                                      & \textbf{51.81}                                                     & \textbf{51.56}                                                           & 280                              & 88.64          & \textbf{57.34}                                                  & \textbf{57.05}       \\     
\bottomrule
\end{tabular}}
\vspace{-0.2cm}
\end{table}
\begin{table*}[t]
\caption{\textbf{CIFAR-100, ImageNette:} Performance (\%) of the proposed defense DAJAT when compared to the state-of-the-art. Robust evaluations are performed on GAMA \cite{sriramanan2020gama} and AutoAttack \cite{croce2020reliable}.}
\vspace{-0.2cm}
\setlength\tabcolsep{2pt}
\resizebox{1.0\linewidth}{!}{
\label{table:sota_c100}
\begin{tabular}{l|c|c|c|c||c|c|c||c|c|c}
\toprule
               & \multicolumn{1}{c|}{}       & \multicolumn{3}{c||}{\textbf{CIFAR-100, ResNet-18}} & \multicolumn{3}{c||}{\textbf{CIFAR-100, WideResNet-34}} & \multicolumn{3}{c}{\textbf{IN-10, ResNet-18}} \\
                       \cmidrule{3-11}
                        \textbf{Method} & \textbf{No. Steps} & \textbf{~Clean~}        & \textbf{~GAMA~}        & \textbf{AutoAttack}        & \textbf{~Clean~}          & \textbf{~GAMA~}         & \textbf{AutoAttack}         & \textbf{~Clean~}       & \textbf{~GAMA~}       & \textbf{AutoAttack}      \\
                        \midrule
 TRADES-AWP \cite{zhang2019theoretically,wu2020adversarial}& 10             & 58.81                 & 25.51                & 25.30                      & 62.41                   & 29.70                 & 29.54                       & 82.73                & 57.52               & 57.40                    \\
TRADES-AWP-WA  & 10          & 59.88                 & 25.81                & 25.52                      & 62.73                   & 29.92                 & 29.59                       & 82.03                & 57.04               & 56.89                    \\
ACAT, Ours (Base, 2step)  & 2     & 62.05                 & 26.35                & 26.10                      & 65.75                   & 30.61                 & 30.23                       & 82.34                & 57.12               & 56.96                    \\
DAJAT, Ours (Base, AA)  & 2 + 2   & 65.75                 & 27.58                & 27.21                      & 67.82                   & \textbf{31.65}                 & 31.26                       & 85.27                & 61.50               & 61.19                    \\
DAJAT, Ours (Base, 2*AA)  & 2 + 4  & 66.84                 & 27.61                & 27.32                      & 68.74                   & 31.58                 & \textbf{31.30}                       & 86.01                & \textbf{62.52}               & \textbf{62.31}                    \\
DAJAT, Ours (Base, 3*AA) & 2 + 6   & \textbf{66.96}                 & \textbf{27.90}                &\textbf{ 27.62 }                     & \textbf{70.35}                   & 31.15                 & 30.89                       & \textbf{86.92}                & 62.14               & 61.89               \\
\bottomrule
\end{tabular}}
\vspace{-0.4cm}
\end{table*}
\begin{table*}[t]
\begin{minipage}{0.48\linewidth}
\caption{Performance (\%) of DAJAT using \textbf{higher attack steps} and 200 training epochs on \textbf{CIFAR-10 and CIFAR-100}. Evaluations are performed against AutoAttack \cite{croce2020reliable} and Common Corruptions (CC) \cite{hendrycks2019benchmarking}.}
\setlength\tabcolsep{2pt}
\resizebox{1.0\linewidth}{!}{
\label{table:200_epochs}
\begin{tabular}{l|c|c|c|c|c}
\toprule
\textbf{Dataset}                                               & \textbf{Model}                       &
\multicolumn{1}{c|}{\textbf{Method}}     & \textbf{Clean} & \textbf{AutoAttack} & \textbf{CC}                   \\
\midrule
                                    &                                      & TRADES-AWP-WA & 81.99 & 51.45               & 72.64      \\
                                    & \multirow{-2}{*}{ResNet-18} & Ours (Base, 2*AA )     & \textbf{85.71}          & \textbf{52.50}               & \textbf{76.13}      \\
                                     \cmidrule{2-6}
                                    &                                                              & TRADES-AWP-WA          & 85.36          & 56.17               & 75.83 \\
                                    
\multirow{-4}{*}{CIFAR-10} & \multirow{-2}{*}{WRN-34-10}                           & Ours (Base, 2*AA )     & \textbf{88.71}          & \textbf{57.81}               & \textbf{80.12}                         \\
\cmidrule{1-6}
                                    &                                                              & TRADES-AWP-WA          & 59.11                                  & 25.97                                       &   47.95    \\
                                     
                                    & \multirow{-2}{*}{ResNet-18}                                  & Ours (Base, 2*AA )     & \textbf{65.45}                                  & \textbf{27.69}               &  \textbf{54.85}     \\
                                     \cmidrule{2-6}
                                   
                                    &                                                              & TRADES-AWP-WA          & 60.30          & 28.68               & 48.95 \\
                                    
\multirow{-4}{*}{CIFAR-100}         & \multirow{-2}{*}{WRN-34-10}                           & Ours (Base, 2*AA )     & \textbf{68.75}                                  & \textbf{31.85}               & \textbf{56.95}  \\

\bottomrule
\end{tabular}
}
\vspace{-0.5cm}
\end{minipage}
\hfill
\begin{minipage}{0.48\linewidth}
\centering
\includegraphics[width=\linewidth]{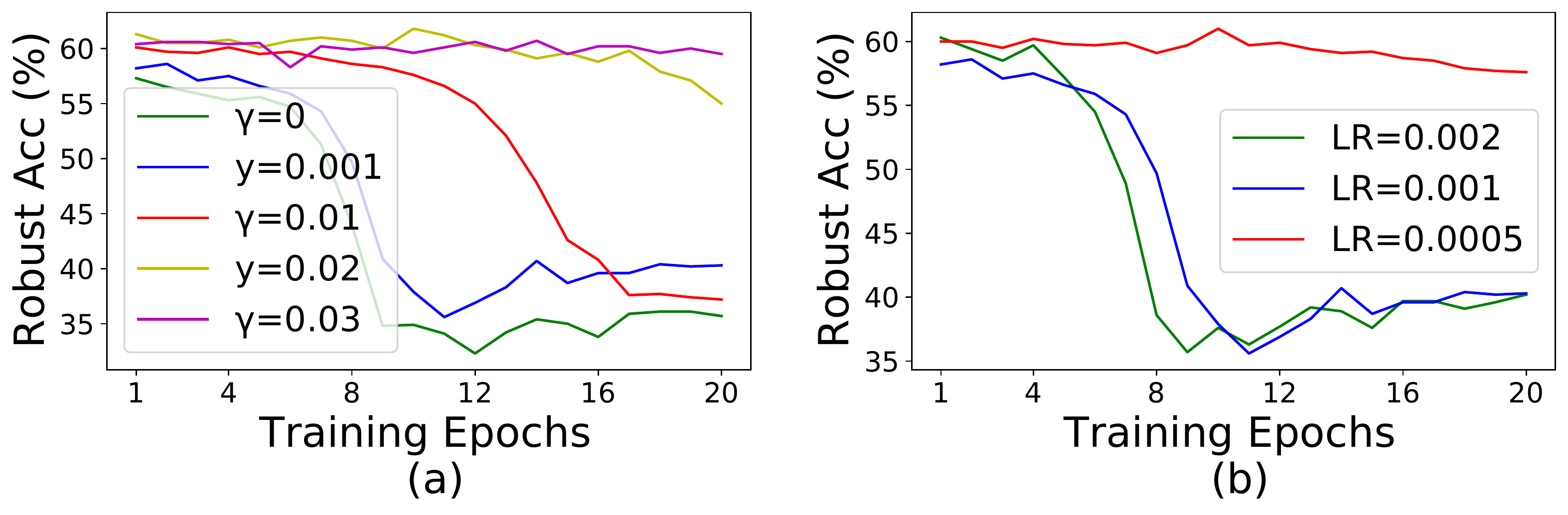}
 \vspace{-0.5cm}
\captionof{figure}{\textbf{Impact of (a) AWP step size $\gamma$ and (b) learning rate (LR) on stability of single-step TRADES-AWP training} using $\varepsilon=7.5/255$ on a model pretrained upto $\varepsilon=7.25/255$ using ACAT. (a) For a fixed LR, increasing $\gamma$ improves training stability. (b) For a fixed value of $\gamma=0.01$, lower LR leads to better stability.}
\vspace{-0.7cm}
\label{fig:gamma}
\end{minipage}
\end{table*}

\vspace{-0.4cm}
\textbf{ACAT:} We compare the performance of the proposed 2-step defense ACAT with the existing state-of-the-art 2-step defense NuAT2-WA \cite{sriramanan2021nuat} in the first partition of Table-\ref{table:sota_c10}. While we achieve similar performance on ResNet-18, we obtain improvements in both clean and robust accuracy on WideResNet-34-10. We show in Section-\ref{subsec:nuat_gat_supple} that our proposed defense ACAT can be integrated with the Nuclear Norm training objective as well to obtain improved results. The performance of the proposed ACAT defense is superior to PGD-AT (10-step) defense \cite{rice2020overfitting} as well. When compared to the TRADES-AWP (10-step) defense, ACAT achieves substantial gains in clean accuracy and improved robust accuracy on CIFAR-100 (Table-\ref{table:sota_c100}), and improved clean accuracy at a slight drop in robust accuracy on CIFAR-10 (Table-\ref{table:sota_c10}), at half the computational cost.

\textbf{DAJAT:} We present three variants of the proposed defense DAJAT, by using one, two and three AutoAugment based augmentations for every image, denoted as (Base, AA), (Base, 2*AA) and (Base, 3*AA) respectively. From Tables-\ref{table:sota_c10} and \ref{table:sota_c100}, we note that even by using only a single augmentation (Base, AA), we obtain improved clean and robust accuracy when compared to most of the baselines considered across all datasets and models. By increasing the number of augmentations to 2 (Base, 2*AA), we observe consistent gains in robust and clean accuracy in all cases. In this setting, the computational complexity of the proposed approach matches with that of TRADES-AWP \cite{wu2020adversarial} as shown in Table-\ref{table:sota_c10}. With the setting (Base, 3*AA), we further obtain marginal improvements in performance over (Base, 2*AA). Overall, using the (Base, 2*AA) approach, which has comparable time complexity as the TRADES-AWP 10-step defense, we obtain large gains ranging from 3.8\% to 7\% on clean accuracy and $\sim1.8\%$ in robust accuracy against AutoAttack \cite{croce2020reliable} across most settings. On the Imagenette dataset we obtain 4.2\% higher clean accuracy and 4.49\% higher robust accuracy, showing that augmentation strategies work best when the amount of training data is less when compared to the complexity of the task. We observe a similar trend using smaller subsets of training data on CIFAR-10 in Fig. \ref{fig:vary_data}.

\textbf{Scaling to higher attack steps, longer training and larger models:} 
While the use of 2 attack steps helps in improving the training efficiency of DAJAT, we show that by using more attack steps and longer training epochs, we can indeed obtain further gains in performance. The varying $\varepsilon$ schedule in DAJAT allows the use of an increasing schedule in the number of steps as well, thereby limiting the overall cost associated with higher attack steps. We present results by increasing the number of attack steps from 2 to 5 uniformly every 50 epochs in Table-\ref{table:200_epochs}. 
We obtain $\sim3.5\%$ and $6-9\%$ higher clean accuracy, along with $1-1.6\%$ and $1.7-3.2\%$ gains in robust accuracy against AutoAttack on CIFAR-10 and CIFAR-100 respectively when compared to the TRADES-AWP-WA 200 epoch baseline. This achieves better robustness-accuracy trade-off when compared to all existing defenses that use the same model architecture on the RobustBench leaderboard \cite{croce2021robustbench}. On the CIFAR-100 RobustBench leaderboard, we outperform all existing methods using the same model architecture, w.r.t. both clean and robust accuracy, including the ones that use additional data, showing that data augmentations can be effectively used to overcome the large sample complexity of adversarial training. The use of diverse augmentations in DAJAT also improves the generalization of the model to Common Corruptions by 4\% and 8\% on CIFAR-10-C and CIFAR-100-C datasets \cite{hendrycks2019benchmarking} respectively as shown in Table-\ref{table:200_epochs}. Further, we find that the performance gains obtained using DAJAT are higher on larger capacity models (Ref:Table-\ref{table:SOTA_WRN3420}).

\textbf{Computational Efficiency:} As shown in Appendix-\ref{app:flops}, the use of ACAT strategy in the proposed DAJAT defense enables us to achieve similar computational complexity as TRADES-AWP defense, while obtaining gains in performance. Using DAJAT, we achieve 10\% reduction in FLOPs (training) and training time, while obtaining 3.8-5.5\% higher clean accuracy and 1-1.6\% higher robust accuracy.

\vspace{-0.35cm}

\subsection{Ablation experiments}
\vspace{-0.35cm}

We present ablation experiments to highlight the significance of different components of the proposed approach in Table-\ref{table:ablation} on the CIFAR-10 dataset using ResNet-18 architecture. All experiments are run for 110 training epochs, except A7 which is run for 220 epochs. 
We show the importance of the JS divergence term in the proposed loss in the ablations A1-A6 of Table-\ref{table:ablation}. Using the JS divergence term we obtain $\sim 1\%$ higher clean accuracy across (Base, AA), (Base, 2*AA) and (Base, 3*AA) settings of the proposed defense. For (Base, 2*AA) and (Base, 3*AA) we obtain marginal improvements in robust accuracy as well. From A7, A9 and A10, we find that the proposed JS divergence term helps even in the case where both augmentations of an image are generated using the same pipeline. Using two AutoAugment based transformations, we obtain 1.6\% higher clean accuracy when compared to the 220 epoch 2-step defense at a comparable computational cost. Comparing A9, A10 and A11, we note that the use of simple and complex augmentations indeed shows improvements over the case of using 2 complex or simple augmentations alone. The importance of split-batch norm in the proposed approach can be evidently seen by comparing A12 and A14. By using single batch norm (A12), robust accuracy drops by 8.24\%. Further, in this case if the JS term is also dropped, the robustness of the network is almost completely lost. This shows that using a single batch norm layer for diverse augmentations makes it harder for the network to converge. We also note that the JS divergence term indeed helps in improving the convergence of training in addition to improving performance. From Appendix-\ref{app:dajat-fixed}, we note that the use of varying $\varepsilon$ schedule in DAJAT improves the stability of training resulting in large performance gains w.r.t. the use of fixed $\varepsilon$, specifically at larger model capacities.

\begin{table}
\caption{\textbf{Ablation experiments} performed on the \textbf{CIFAR-10} dataset using ResNet-18 architecture. Robust Accuracy is reported against GAMA attack \cite{sriramanan2020gama} }
\setlength\tabcolsep{2pt}
\resizebox{1.0\linewidth}{!}{
\label{table:ablation}
\begin{tabular}{l|c|c|c||l|c|c|c}
\toprule

\multicolumn{1}{l|}{\textbf{Method}}                   & \textbf{\# Steps}                            & \textbf{~Clean~}            & \textbf{Robust}           & \multicolumn{1}{l|}{\textbf{Method}}              & \textbf{\# Steps}             & \textbf{~Clean~} & \textbf{Robust}  \\
\midrule
 
{[}A1{]} Ours (Base, AA ), no JS div                          & 2 + 2                                        & 84.55                         & 51.45                         & {[}A8{]} Ours (Base, 2step)                    & 2                             & 82.41              & 50.00               \\
 
{[}A2{]} Ours (Base, AA )                                      & 2 + 2                                        & 85.60                         & 51.27                         & {[}A9{]} Ours (AA, AA )                                   & 2 + 2                         & 84.68              & 49.71               \\
{[}A3{]} Ours (Base, 2*AA ), no JS div & 2 + 4                & 85.07 & 51.53 & {[}A10{]} Ours (Base, Base )      & 2 + 2 & 83.93              & 49.88               \\
 
{[}A4{]} Ours (Base, 2*AA )                                    & 2 + 4                                        & 85.99                         & 51.71                         & {[}A11{]} Ours (Base, AA )                                & 2 + 2                         & 85.60              & 51.27               \\
 
{[}A5{]} Ours (Base, 3*AA ), no JS div                        & 2 + 6                                        & 85.31                         & 51.67                         & {[}A12{]} Ours (Base, 3*AA ) Single BN            & 2 + 6                         & 86.68              & 43.57               \\
 
{[}A6{]} Ours (Base, 3*AA )                                    & 2 + 6                                        & 86.67                       & 51.81                        & {[}A13{]} Ours (Base, 3*AA ) Single BN, no JS & 2 + 6                         & 75.64              & 4.20                \\
 
{[}A7{]} Ours (Base, 2step), 220 epochs                         & 2                                            & 83.05                         & 50.31                         & {[}A14{]} Ours (Base, 3*AA )                              & 2 + 6                         & 86.67              & 51.81 \\
\bottomrule

\end{tabular}}
\end{table}
\begin{table*}[t]
\caption{Impact of using \textbf{other augmentations} in DAJAT. Performance on CIFAR-10 dataset with ResNet-18 architecture is reported. Robust evaluations are done against GAMA attack \cite{sriramanan2020gama}. $^\dagger$PreAct-ResNet18 with Swish activation is used \cite{rebuffi2021data,rade2021pytorch}.}
\vspace{-0.2cm}
\setlength\tabcolsep{2pt}
\resizebox{1.0\linewidth}{!}{
\label{table:aug}
\begin{tabular}{l|c|c|c|c|c|c||l|c|c|c|c|c|c}
\toprule
\multicolumn{1}{l|}{}  & \multicolumn{2}{c|}{\textbf{Augmentation}} & \multicolumn{2}{c|}{\textbf{Base + Aug }} & \multicolumn{2}{c||}{\textbf{Base + 2 * (Aug) }}  & \multicolumn{1}{l|}{}  & \multicolumn{2}{c|}{\textbf{Augmentation}} & \multicolumn{2}{c|}{\textbf{Base + Aug }} & \multicolumn{2}{c}{\textbf{Base + 2 * (Aug) }} \\
\cmidrule{2-7}
\cmidrule{9-14}
\textbf{Augmentation} & \textbf{Clean}  & \textbf{Robust} & \textbf{Clean}        & \textbf{Robust}       & \textbf{Clean}           & \textbf{Robust} & \textbf{Augmentation} & \textbf{Clean}  & \textbf{Robust} & \textbf{Clean}        & \textbf{Robust}       & \textbf{Clean}           & \textbf{Robust}           \\
\midrule
No Augmentation       & 76.32               & 43.20               & 78.08                     & 41.71                     & 77.42                        & 41.07                        & Cutout   \cite{cutout}             & 82.38               & \textbf{50.14}               & 84.91                     & \textbf{51.40}                     & 85.11                        & 51.60                        \\
Pad+Crop+H-Flip       & 82.41               & 50.00               & 83.69                     & 51.30                     & 83.62                        & 51.09                        &  Colour Jitter           & \textbf{82.98}        & 48.82    & 84.50            & 51.19            & 84.85                                                  & 51.62                                                 \\
AutoAugment  \cite{cubuk2018autoaugment}         & 82.54               & 48.11               & 84.94                     & 51.23                     & \textbf{85.99}                        & \textbf{51.71}                        & Mixup  \cite{zhang2017mixup}               & 79.08               & 45.07               & 85.18                     & 50.18 &  84.24                        & 50.01    \\
Cutmix     \cite{yun2019cutmix}           & 79.03               & 41.57               & 82.33                     & 50.90                     & 81.64                        & 49.50             &   RandAugment  \cite{cubuk2020randaugment}          & 82.48        & 44.66       & 84.61        & 51.01        & 85.47                                                      & 51.33                                                     \\ Cutmix$^\dagger$ \cite{rebuffi2021data,rade2021pytorch}           & 82.01        & 47.65        & 84.58        & 50.97        & 85.49 & 	51.58   &   Augmix    \cite{hendrycks2020augmix}     & 82.38        & 48.84        & \textbf{84.96}        & 50.4        & 85.18 & 	50.51                                                 \\

\bottomrule

\end{tabular}}
\vspace{-0.3cm}
\end{table*}

\vspace{-0.2cm}
\subsection{Combining the proposed defense with other augmentations}
\vspace{-0.2cm}
We explore combining the proposed defense DAJAT with other augmentations in Table-\ref{table:aug}, and the impact of individual augmentations in AutoAugment in Table-\ref{table:augs}. We do not use the JS divergence term for Cutmix and Mixup since they involve changes in the label space. Without using any augmentation in the training dataset, we obtain poor clean and robust accuracy, highlighting the importance of the simple augmentations. The proposed approach is able to obtain good performance gains using AutoAugment \cite{cubuk2018autoaugment}, Color Jitter and CutOut \cite{cutout} augmentations, highlighting that it can work well with pixel-level and spatial augmentations. In case of CutMix \cite{yun2019cutmix} and Mixup \cite{zhang2017mixup}, since the base accuracy obtained by using the augmentation alone is itself very poor, combining them with the proposed approach does not yield competent results. However they are considerably better than naively using the augmentation. This shows that the proposed approach enables the use of a variety of augmentations without the need for careful selection. The use of Cutmix and Mixup in adversarial training are challenging since they involve changes in label space. We note that Rebuffi et al. \cite{rebuffi2021data} obtain considerable gains using Cutmix along with many other improvements. However, naively augmenting using CutMix leads to a considerably degraded performance. By incorporating some of the tricks reproduced by Rade et al. \cite{rade2021pytorch} we obtain improvements in the CutMix baseline and obtain further gains in the proposed method as well (Ref: Appendix-\ref{sec:cutmix_deepmind}, Table-\ref{table:cutmix}).

\vspace{-0.3cm}
\section{Conclusion}
\vspace{-0.2cm}
\label{sec:conclusions}
Contrary to prior knowledge, we show that it is indeed possible to use common augmentation strategies that modify the low-level statistics of images to improve the performance of adversarial training. We propose a novel defense Diverse Augmentation based Joint Adversarial Training (DAJAT), that uses a combination of simple and complex augmentations with separate batch normalization layers to allow the network training to benefit from the diverse training data distribution obtained using complex augmentations, while also specializing on a distribution that is close to the test set. The use of JS divergence term between network predictions of different augmentations enables the joint learning across various augmentations. We improve the efficiency of the proposed defense by utilizing the proposed approach Ascending Constraint Adversarial Training (ACAT), that improves the stability and performance of TRADES 2-step adversarial training significantly by using a linearly increasing $\varepsilon$ schedule along with a cosine learning rate schedule and weight-space smoothing. \\
\textbf{Limitations:} While we focus our work on \textit{how} to use augmentations effectively in adversarial training, we do not focus on \textit{which} augmentations are best suited for the same. We believe this work can open up further possibilities towards finding better data augmentations for adversarial training. 
\vspace{-0.2cm}
\section{Acknowledgments and Disclosure of Funding}
\vspace{-0.2cm}
This work was supported by a research grant (CRG/2021/005925) from SERB, DST, Govt. of India. Sravanti Addepalli is supported by a Google PhD Fellowship in Machine Learning.

\bibliographystyle{abbrvnat}
\small{\bibliography{references}}

\section*{Checklist}

\begin{enumerate}

\item For all authors...
\begin{enumerate}
  \item Do the main claims made in the abstract and introduction accurately reflect the paper's contributions and scope?
    \answerYes
  \item Did you describe the limitations of your work?
    \answerYes{Section-\ref{sec:conclusions}}
  \item Did you discuss any potential negative societal impacts of your work?
    \answerNA{There are no known negative societal impacts of our work.}
  \item Have you read the ethics review guidelines and ensured that your paper conforms to them?
    \answerYes
\end{enumerate}

\item If you are including theoretical results...
\begin{enumerate}
  \item Did you state the full set of assumptions of all theoretical results?
    \answerNA
        \item Did you include complete proofs of all theoretical results?
    \answerNA
\end{enumerate}

\item If you ran experiments...
\begin{enumerate}
  \item Did you include the code, data, and instructions needed to reproduce the main experimental results (either in the supplemental material or as a URL)?
    \answerYes{\url{https://github.com/val-iisc/DAJAT}.}
  \item Did you specify all the training details (e.g., data splits, hyperparameters, how they were chosen)?
    \answerYes{Appendix-\ref{sec:hyp_sens} and ReadMe file at \url{https://github.com/val-iisc/DAJAT}}
        \item Did you report error bars (e.g., with respect to the random seed after running experiments multiple times)?
    \answerYes{Table:\ref{table:variance}}
        \item Did you include the total amount of compute and the type of resources used (e.g., type of GPUs, internal cluster, or cloud provider)?
    \answerYes{Details on compute: Table-\ref{table:flops_WRN}, Details on resources: Appendix- \ref{sec:hyp_sens} }
\end{enumerate}

\item If you are using existing assets (e.g., code, data, models) or curating/releasing new assets...
\begin{enumerate}
  \item If your work uses existing assets, did you cite the creators?
    \answerYes{Appendix- \ref{sec:dataset_details}}
  \item Did you mention the license of the assets?
    \answerYes{Appendix- \ref{sec:dataset_details}}
  \item Did you include any new assets either in the supplemental material or as a URL?
    \answerYes{We share the code for the proposed approach.}
  \item Did you discuss whether and how consent was obtained from people whose data you're using/curating?
    \answerNA{We use only publicly released and open-sourced datasets and codes.}
  \item Did you discuss whether the data you are using/curating contains personally identifiable information or offensive content?
    \answerYes{Appendix-\ref{sec:dataset_details}}
\end{enumerate}

\item If you used crowdsourcing or conducted research with human subjects...
\begin{enumerate}
  \item Did you include the full text of instructions given to participants and screenshots, if applicable?
    \answerNA
  \item Did you describe any potential participant risks, with links to Institutional Review Board (IRB) approvals, if applicable?
    \answerNA
  \item Did you include the estimated hourly wage paid to participants and the total amount spent on participant compensation?
    \answerNA
\end{enumerate}

\end{enumerate}

\clearpage

\appendix

\section{Details on the proposed defense DAJAT}

The algorithm of the proposed approach is presented in Algorithm-\ref{alg:DAJAT}. In every training iteration, multiple augmentations are considered for every image $x_i$ (L7). We consider one base augmentation and $T$ complex augmentations. The base augmentation consists of Pad and Crop followed by Horizontal Flip, while the complex augmentations are a combination of AutoAugment \cite{cubuk2018autoaugment} and the base augmentations. The attack generation for each augmentation (L8-L13) is similar to the ACAT algorithm discussed in Section-\ref{sec:acat_algo}. The DAJAT loss (L16) is a combination of the TRADES loss \cite{zhang2019theoretically} (L17) on each augmentation, and a Jensen-Shannon (JS) divergence term between all augmentations. The JS divergence is a combination of KL divergence terms with respect to the average probability vector as shown below.
\begin{multline}
    \mathrm{JSD}(f_\theta(x_{i,\mathrm{base}}),f_\theta(x_{i,\mathrm{auto(1)}}), \dots , f_\theta(x_{i,\mathrm{auto(T)}}))  = \frac{1}{T+1} \big\{ \mathrm{KL}(f_\theta(x_{i,\mathrm{base}},M)) + \\ \mathrm{KL}(f_\theta(x_{i,\mathrm{auto(1)}},M)) + \dots + \mathrm{KL}(f_\theta(x_{i,\mathrm{auto(T)}},M)) \big\}
\end{multline}

where $M$ is defined as below,
\begin{equation}
    M = \frac{1}{T+1} \big\{f_\theta(x_{i,\mathrm{base}}) + f_\theta(x_{i,\mathrm{auto(1)}}) + \dots + f_\theta(x_{i,\mathrm{auto(T)}})\big\}
\end{equation}
As shown in Table-\ref{table:ablation}, the JS-divergence term improves accuracy on clean samples and training convergence by enabling the joint learning of representations across different augmentations. The model weights are perturbed by $\tilde{\theta}$, by maximizing the TRADES loss on the base augmentations alone within the constraint set $\mathcal{M}(\theta)$ (L18). This constraint set is chosen such that $||\tilde{\theta}_l|| \leq \gamma \cdot ||\theta_l||$ for any layer $l$. The network at $\theta$ is then updated using gradients at $f_{\theta+\tilde{\theta}}$ to minimize the overall loss $\mathcal{L}_\mathrm{DAJAT}(\theta+\tilde{\theta}$) (L20).

\section{Augmentations}
\begin{algorithm}[tb]
   \caption{Diverse Augmentation based Joint Adversarial Training (DAJAT)}
   \label{alg:DAJAT}
   
\begin{algorithmic}[1]
   \STATE {\bfseries Input:} Network $f_\theta$, Training Dataset $\mathcal{D} =\{(x_i,y_i) \}$, Adversarial Threat model: $\ell_\infty$ bound of radius $\varepsilon$, number of epochs E, Maximum Learning Rate $\mathrm{LR}_{max}$, $M$ training mini-batches of size $n$, number of attack steps $S$, Cross-entropy loss $\ell_{CE}$, Weight perturbation constraint $\mathcal{M}(\theta)$, Number of augmented images using autoaugment $T$, coefficient of KL divergence term $\beta$
   
\FOR{$epoch=1$ {\bfseries to} $E$}
    \STATE $\varepsilon_{\mathrm{asc}} = epoch \cdot \varepsilon/ E $
    \STATE $\mathrm{LR} = 0.5  \cdot \mathrm{LR}_{max} \cdot (1 + cosine((epoch -1)/E\cdot \pi))$
    \FOR{$iter=1$ {\bfseries to} $M$}
        \FOR{$i=1$ {\bfseries to} $n$ (in parallel)} 
        \FOR{$a \in \{\mathrm{base},\mathrm{auto(1)},\dots ,\mathrm{auto(T)}\}$}
        \FOR{$steps=1$ {\bfseries to} $S$}
        \STATE $\delta = 0.001 \cdot \mathcal{N}(0,1)$
        \STATE $\delta = \delta + \varepsilon_{\mathrm{asc}} \cdot \mathrm{sign}\left(\nabla_{\delta} \mathrm{KL}(f_{{\theta}}(x_{i,a})||f_{{\theta}}(x_{i,a}+\delta)) \right)$
        \STATE $\delta =Clamp~(\delta,-\varepsilon_{\mathrm{asc}},\varepsilon_{\mathrm{asc}})$
        \STATE $\widetilde{x}_{i,a} = Clamp~(x_{i,a} + \delta,0,1)$
        \ENDFOR
        \ENDFOR
        \ENDFOR
        \STATE $\begin{aligned}
   \mathcal{L}_{\mathrm{DAJAT}}(\theta) = \frac{1}{T+1} \cdot \frac{1}{n} \sum\limits_{i=1}^n \big\{ \mathcal{L}_{\mathrm{TR}}({\theta}, (x_i,\tilde{x}_i)_\mathrm{base},y_i) +  \sum\limits_{t=1}^T  \mathcal{L}_{\mathrm{TR}}(\theta, (x_i,\tilde{x}_i)_\mathrm{auto},y_i)\big\} \\ +    \frac{1}{n} \sum\limits_{i=1}^n \big\{\mathrm{JSD}(f_\theta(x_{i,\mathrm{base}}),f_\theta(x_{i,\mathrm{auto(1)}}), \dots , f_\theta(x_{i,\mathrm{auto(T)}}))\big\}
\end{aligned}$
\vspace{0.2cm}
        \STATE where, $\mathcal{L}_{\mathrm{TR}}(\theta,(x,\tilde{x}),y) =  \mathcal{L}_\mathrm{CE}(f_{\theta}(x),y) + \beta \cdot\mathrm{KL}(f_{\theta}(x)||f_{\theta}(\tilde{x}))$
        
         \STATE $\tilde{\theta} = \argmax\limits_{{\hat{\theta}} \in \mathcal{M}(\theta)} \frac{1}{n} \sum\limits_{i=1}^n \big\{ \mathcal{L}_{\mathrm{TR}}({\theta+\hat{\theta}},(x_i,\tilde{x}_i)_\mathrm{base},y_i)\big\}$
         \vspace{0.1cm}
         \STATE $\theta = \theta -  \mathrm{LR} \cdot \nabla_{\theta} ( \mathcal{L}_{\mathrm{DAJAT}}(\theta+\tilde{\theta}))$
        \ENDFOR
        
\ENDFOR
\end{algorithmic}
\end{algorithm}

\label{app:simplecomplexaugs}

While existing works hypothesize that augmentations changing the low-level statistics of images cannot effectively improve adversarial training \cite{rebuffi2021data}, we show in this work that with the use of split Batch-normalization layers and JS divergence term between different augmentations, it is indeed possible to obtain significant gains using augmentations that modify the low-level statistics of images as well. In this work, we use an existing augmentation strategy, AutoAugment to obtain an improvement in performance using the proposed training algorithm DAJAT. AutoAugment uses Proximal Policy Optimization to find the set of policies that can yield optimal performance on a given dataset. It consists of 25 unique sub-policies for a given dataset, where each sub-policy is a combination of two augmentations chosen from a set of pre-defined augmentations in series. The pre-defined augmentations include the spatial transformations - shear, rotation and translation, and augmentations that cause changes in low-level statistics of images - color, posterize, solarize, brightness, contrast, sharpness, autocontrast, equalize and invert.

\begin{figure*}[h]
\begin{minipage}{0.47\linewidth}
\centering
        \includegraphics[width=\linewidth]{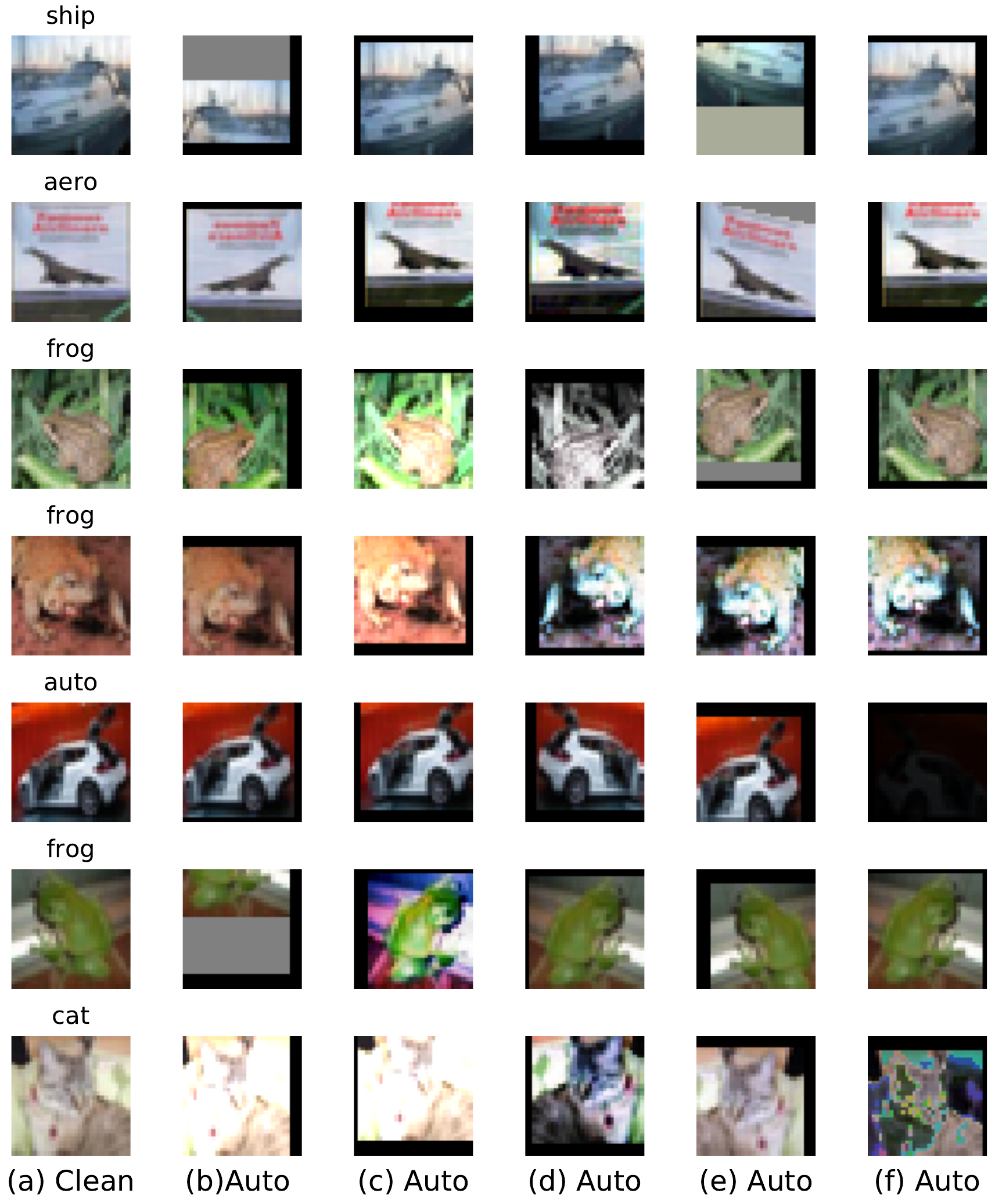}
        \caption{CIFAR-10 images in (a), along with respective random augmentations generated using AutoAugment \cite{cubuk2018autoaugment} shown in columns (b-f)}
        \label{fig:auto}
        \end{minipage}
        \hfill
\begin{minipage}{0.47\linewidth}
\centering
        \includegraphics[width=\linewidth]{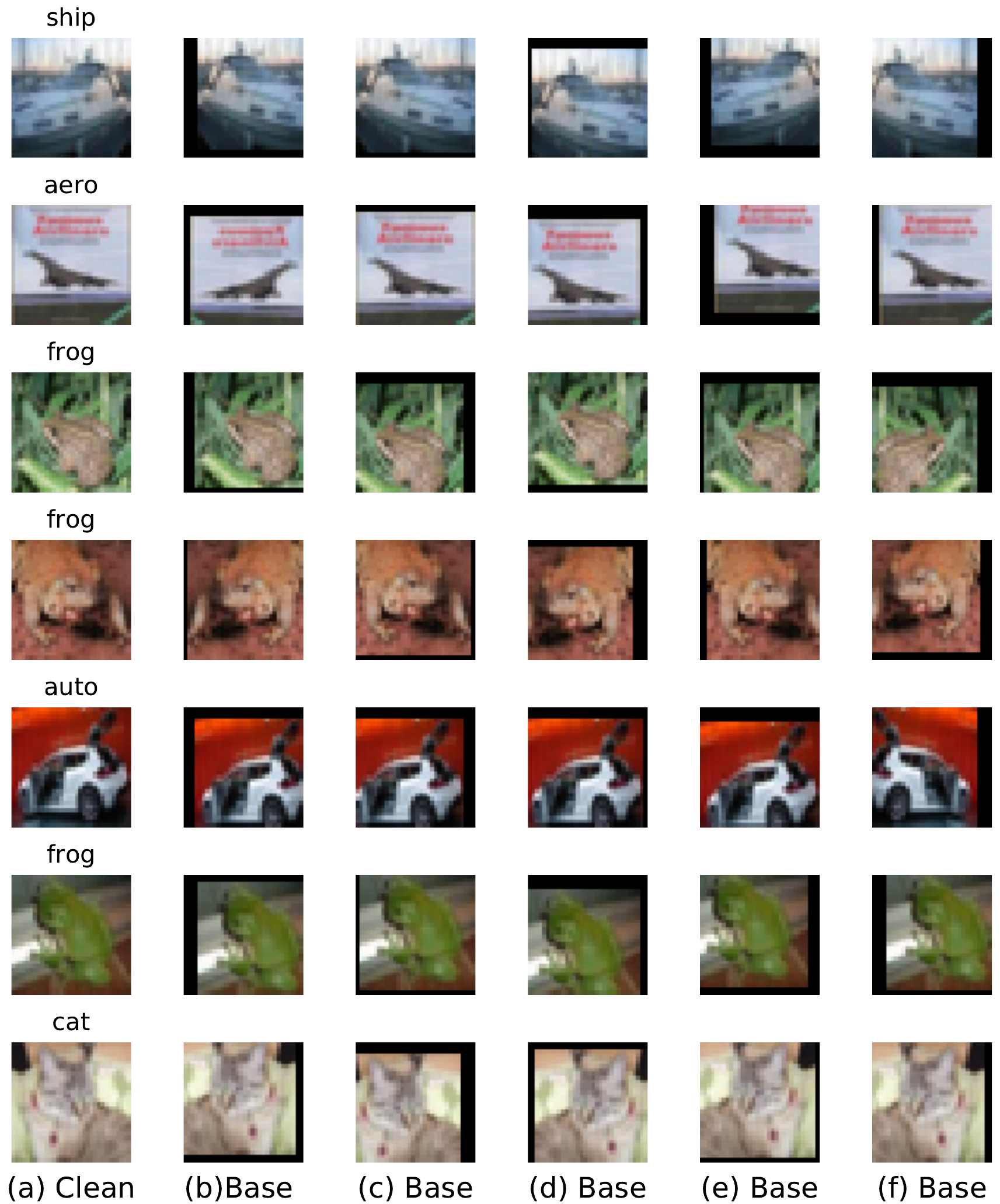}
         \vspace{-0.4cm}
        \caption{CIFAR-10 images in (a), along with respective random augmentations generated using Base augmentations (Pad+Crop and Horizontal Flip) shown in columns (b-f)}
        \label{fig:base}
\end{minipage}
\end{figure*}

\begin{figure}
\centering
\begin{minipage}[h]{0.3\linewidth}
\centering
        \includegraphics[width=\linewidth]{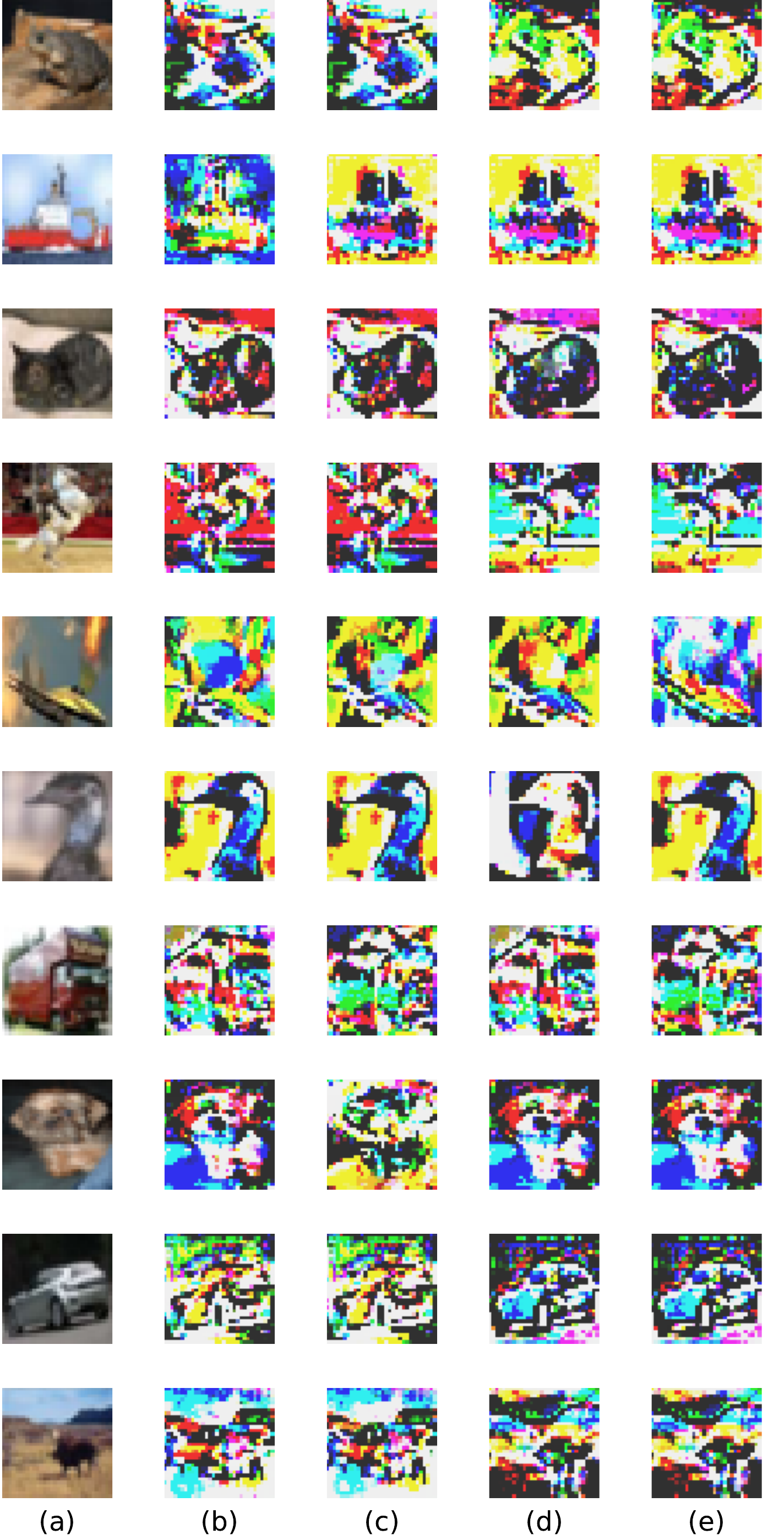}
        \caption{CIFAR-10 images in (a), along with perturbations of respective images generated using without any augmentation shown in columns (b-e). The attack is generated at $\varepsilon=8/255$ and the corresponding perturbation's magnitude is scaled up for better visibility}
        \label{fig:noaug}
        \end{minipage}
        \hfill
\begin{minipage}[h]{0.3\linewidth}
\centering
        \includegraphics[width=\linewidth]{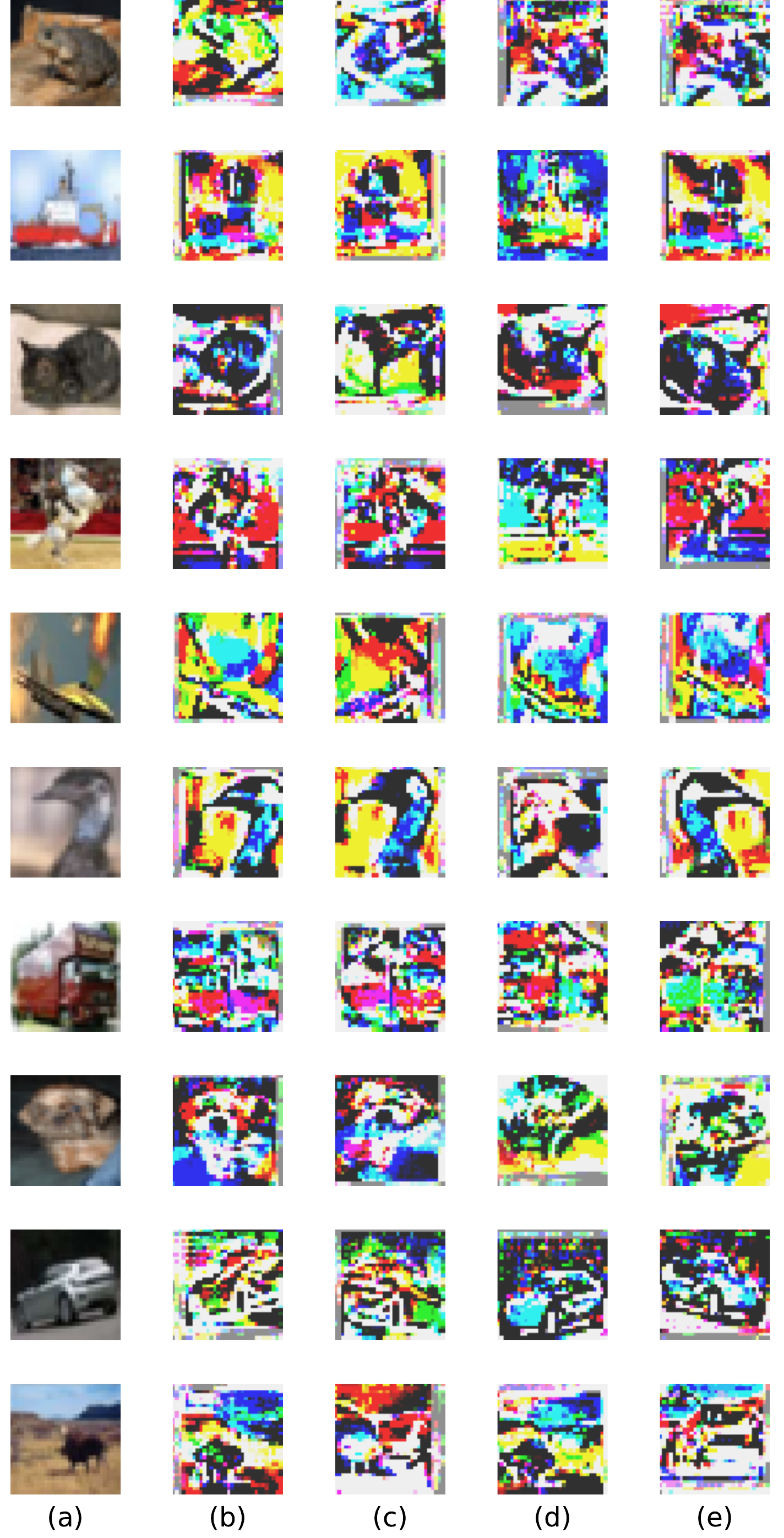}
        \caption{CIFAR-10 images in (a), along with perturbations of respective random augmentations generated using Base augmentations (Pad+Crop and Horizontal Flip) shown in columns (b-e). The attack is generated at $\varepsilon=8/255$ and the corresponding perturbation's magnitude is scaled up for better visibility}
        \label{fig:baseaug}
\end{minipage}
\hfill
\begin{minipage}[h]{0.3\linewidth}
\centering
        \includegraphics[width=\linewidth]{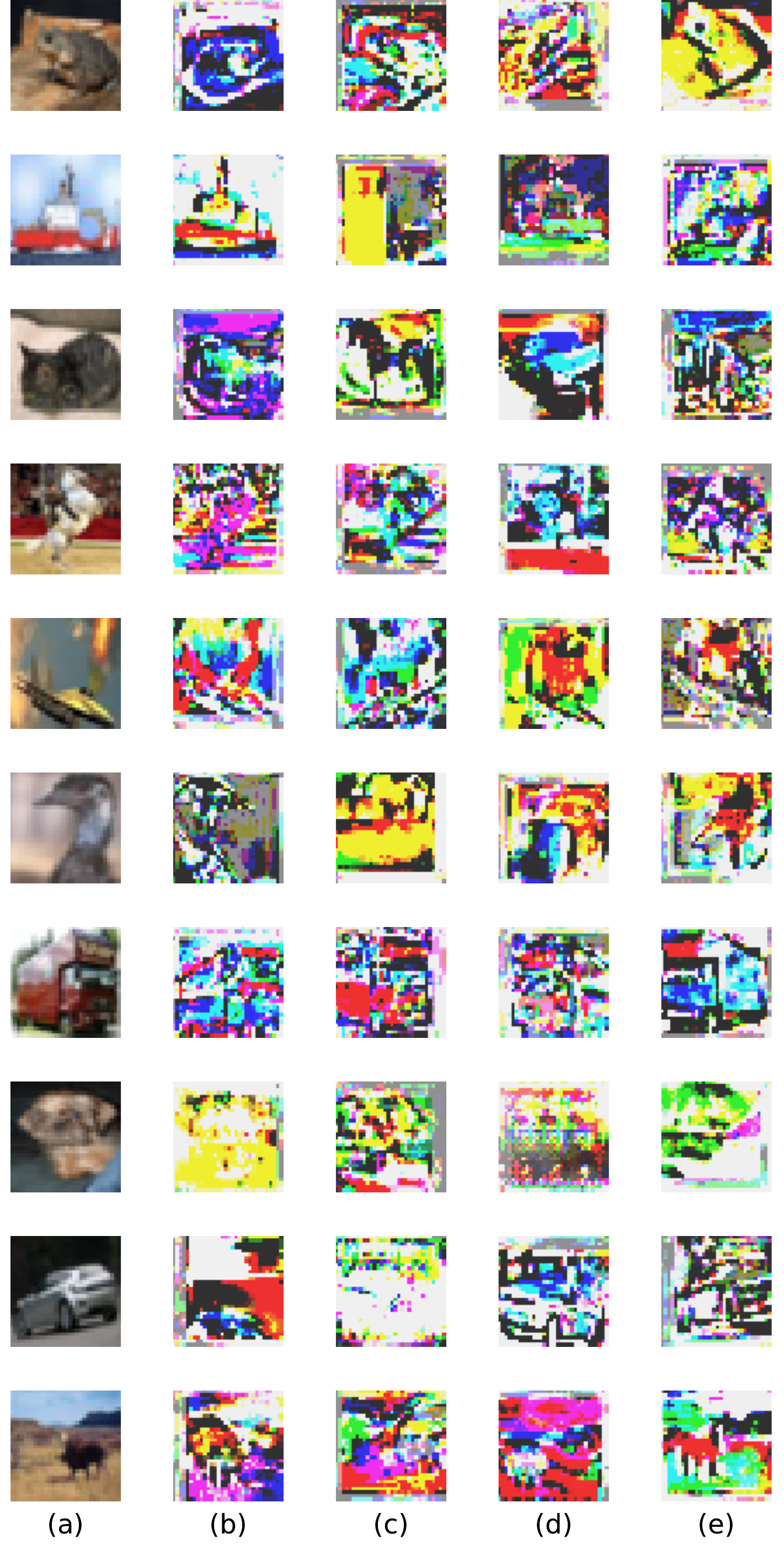}
        \caption{CIFAR-10 images in (a), along with perturbations of respective random augmentations generated using AutoAugment \cite{cubuk2018autoaugment} shown in columns (b-e).The attack is generated at $\varepsilon=8/255$ and the corresponding perturbation's magnitude is scaled up for better visibility}
        \label{fig:autoaug}
\end{minipage}
\vspace{-0.5cm}
\end{figure}

We visualize some of the augmentations generated using AutoAugment and Base augmentations (Pad+Crop, Horizontal Flip) in Fig.\ref{fig:auto} and Fig.\ref{fig:base} respectively. It can be noted that multiple augmentations generated using AutoAugment are significantly more diverse than the augmentations generated using the base augmentations. The use of multiple diverse augmentations leads to improved generalization on test set as discussed in Sec.\ref{sec:motivation}. Further we visualize the perturbations generated using the 2-step attack with KL-divergence loss on CIFAR-10 images without augmentations, with Base augmentations and with AutoAugment on Trades-AWP \cite{wu2020adversarial} model trained at $\varepsilon=8/255$ in Fig.\ref{fig:noaug}, Fig.\ref{fig:baseaug} and Fig.\ref{fig:autoaug} respectively. For plotting, one image is selected at random from each of the ten classes. Because of the increased diversity amongst the images on using AutoAugment, we observe more diversity in the perturbations as well (Fig.\ref{fig:autoaug}) when compared to the Base augmentations (Pad+Crop, Horizontal Flip) (Fig.\ref{fig:baseaug}) and No augmentations (Fig.\ref{fig:noaug}) case.

\subsection{Distinction between Simple and Complex Augmentations}
We term the augmentations that preserve low-level features of images as simple augmentations, and those that modify the same as complex augmentations. To distinguish between simple and complex augmentations, we do not use the difference between two images in pixel-space, since this would incorrectly show that simple changes like horizontal-flip and crop are far apart. Instead, we use metrics that better capture low-level features at pixel and patch levels. This can be measured at a pixel-level using MSE between color histograms, and at a patch-level using patch-wise MSE. To compute patch-wise MSE between two images $x_{1}$ and $x_{2}$ , for every $8\times8$  patch in $x_{1}$ we find the nearest patch in $x_{2}$ and a horizontal flip of $x_{2}$ , and compute an average across all patches in $x_{1}$. We report the mean and standard deviation of this value across all images in the test set. We show the pair-wise distances between three sets of images (Unaugmented, Pad+Crop+HFlip, AutoAugment) in Table-\ref{table:mse_pathc_hist}. Lower value indicates that the images are more similar. We note that Pad+Crop+HFlip augmentations have the advantage of being more similar to the distribution of unaugmented images that are expected during inference, while AutoAugment transformed images are farther away from the unaugmented images. AutoAugment consists of several augmentations of varying complexity levels, and may contain augmentations of similar complexity as the Base augmentations (Pad+Crop+HFlip) as well. This is reflected in the higher variance in pair-wise distances corresponding to AutoAugment.

\begin{table}[h]
\centering
\caption{Distinction between Simple and Complex Augmentations in pixel space, in terms of MSE between Histograms and MSE between Patches.}
\vspace{0.1cm}
\resizebox{1.0\linewidth}{!}{
\label{table:mse_pathc_hist}
\begin{tabular}{l|c|c|c}
\toprule
\multicolumn{1}{c|}{\textbf{Image pairs}}      & \multicolumn{1}{c|}{\textbf{ Complexity}}            & \multicolumn{1}{c|}{\textbf{~~~~MSE between Histograms~~~~}} & \multicolumn{1}{c|}{\textbf{~~MSE between Patches~~~~}} \\ 
\midrule
\multicolumn{1}{c|}{Base (Pad+Crop+HFlip), Unaugmented} & Simple & \multicolumn{1}{c|}{133.60 \std{94.05}}                 & \multicolumn{1}{c|}{43.68 \std{23.37}}               \\
AutoAugment,  Unaugmented         & Complex                       & 289.25 \std{405.11}                                     & 51.39 \std{24.23}     \\
\bottomrule
\end{tabular}}
\end{table}

DAJAT allows separate function mappings for augmentations that resemble the inference-time distribution (Pad+Crop+HFlip), and those that lead to better diversity (AutoAugment). Base augmentations have low variation, and are similar to the distribution of unaugmented images, which is important to obtain performance gains using the batch-norm layer corresponding to these base augmentations during inference. On the other hand, the high variance of AutoAugment based transformations helps in improving the robust generalization of the overall model.

\subsection{Distribution shift of Natural and Adversarial images due to Augmentations}
We compare the distribution shift between (natural-augmented, natural-unaugmented) pairs and (adversarial-augmented, adversarial-unaugmented) pairs. We consider two types of distances between image pairs: low-level (MSE between histograms/patches) and feature-level (FID). In terms of low-level distances, we expect the distances between (natural-augmented, natural-unaugmented) pairs and the corresponding (adversarial-augmented, adversarial-unaugmented) pairs to be similar, since the perturbed images are only an $\varepsilon$ away from natural images. However, as seen in Fig.\ref{fig:noaug}, \ref{fig:baseaug} and \ref{fig:autoaug}, the perturbations of Pad+Crop+HFlip look similar to the perturbations of unaugmented images, while the perturbations of AutoAugment based images look different from those of unaugmented images. This is a result of larger pixel-level differences between the (natural-AutoAugment, natural-unaugmented) image pairs when compared to (natural-PadCrop, natural-unaugmented) image pairs, which serves as a more diverse initialization for the attack. The difference in the absolute perturbations results in a larger distance in feature space (Fréchet Inception Distance or FID) as shown in Table-\ref{table:fid}. 

\begin{table}[h]
\centering
\caption{FID between Augmented and Unaugmented images with Simple and Complex augmentations}
\vspace{0.1cm}
\resizebox{1.0\linewidth}{!}{
\label{table:fid}
\begin{tabular}{l|c|c}
\toprule
\multicolumn{1}{c|}{\textbf{Image pairs}}                & \multicolumn{1}{c|}{\textbf{FID between Natural  image pairs}} & \textbf{FID between Adversarial image pairs} \\
\midrule
\multicolumn{1}{c|}{Base (Pad+Crop+HFlip), Unaugmented} & \multicolumn{1}{c|}{24.02}                                    & 33.41                                        \\ 
AutoAugment,  Unaugmented                                & 37.62                                                         & 43.75     \\
\bottomrule
\end{tabular}}
\end{table}

For better clarity, we summarize our findings in Table-\ref{table:fid_Summary}. The higher feature level distance between (adversarial-AutoAugment, adversarial-unaugmented) image pairs when compared to (natural-AutoAugment, natural-unaugmented) image pairs translates to higher $\frac{1}{2}d_{\mathcal{F}\Delta\mathcal{F}}(s,t)$ in Eq.\ref{bendavid}. Based on Conjecture-1(ii), unless this difference is accounted for, complex augmentations cannot improve the performance of adversarial training.

\begin{table}[]
\centering
\caption{Summary of pixel-level and feature-level distances between Augmented and Unaugmented image pairs for Natural and Adversarial Images. Base refers to the augmentations Pad+Crop+HFlip.}
\vspace{0.1cm}
\resizebox{1.0\linewidth}{!}{
\label{table:fid_Summary}
\begin{tabular}{l|l|c|c}
\toprule
\multicolumn{1}{c|}{\textbf{Natural/ Adversarial}}   & \multicolumn{1}{l|}{\textbf{Distributions}}   & \multicolumn{1}{c|}{\textbf{Low-level distance}} & \multicolumn{1}{c}{\textbf{Feature-level distance}} \\ 
\midrule
\multicolumn{1}{l|}{\multirow{2}{*}{Natural images}} & \multicolumn{1}{l|}{Base, Unaugmented} & \multicolumn{1}{c|}{Low}                         & Low                             \\ 
\multicolumn{1}{c|}{}                                & Autoaugment, Unaugmented                   & High                                             & Medium                          \\ 
\midrule
\multirow{2}{*}{Adversarial images}                    & Base, Unaugmented                      & Low                                              & Medium                          \\
                                                      & Autoaugment, Unaugmented                   & High                                             & \textbf{High}                   \\
\bottomrule
\end{tabular}}
\vspace{-0.1cm}
\end{table}

\subsection{Justification on the choice of Simple and Complex Augmentation pipeline}
\label{appsubsec:justifyaugs}
\begin{figure}[]

\centering
        \includegraphics[width=\linewidth]{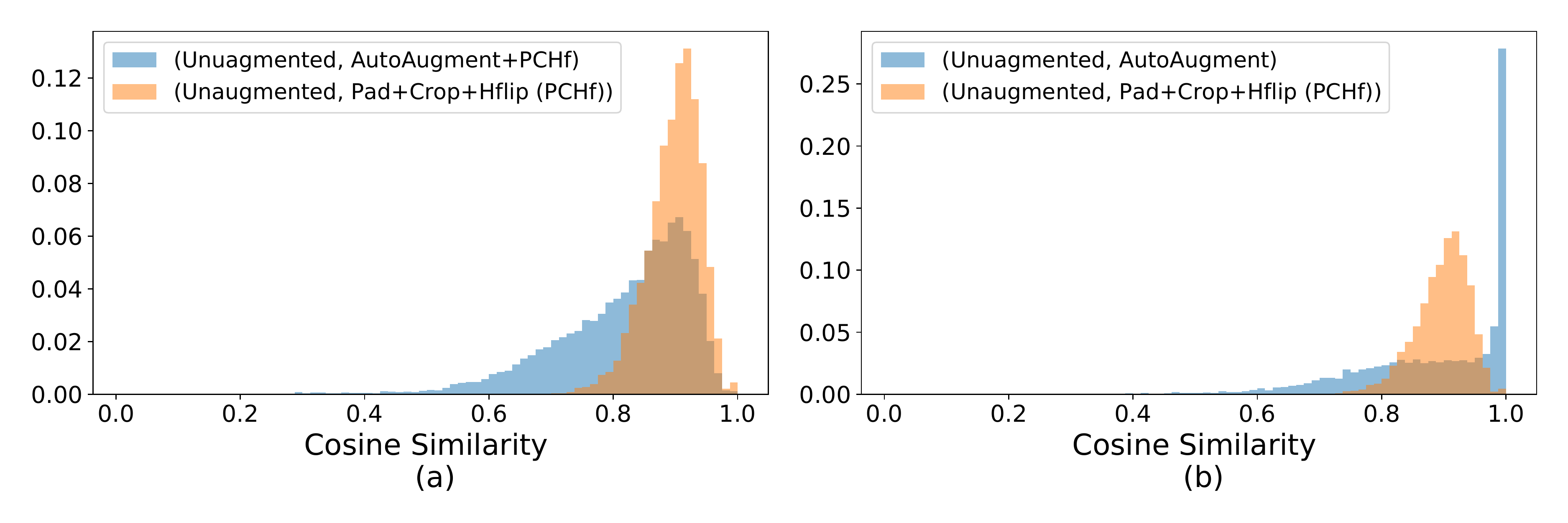}
        \caption{Normalized histogram of pair-wise cosine similarity between the features obtained using a pretrained Inception-V3 \cite{Szegedy2016RethinkingTI} model for different augmentations w.r.t. the respective input images. Histograms are plotted for the following pairs, (a) (Unaugmented, AutoAugment+Pad+Crop+Horizontal flip) and (Unaugmented, Pad+Crop+Horizontal flip), (b) (Unaugmented, AutoAugment) and (Unaugmented, Pad+Crop+Horizontal flip). The full test set of CIFAR-10 (10k images) is used for the plots.}
        \label{fig:aa_pc_hist}
\end{figure}

In the proposed method, we use two sets of augmentations - Pad+Crop+HFlip and AutoAugment. The second set of augmentations (complex augmentations) consists of an autoaugment based transformation followed by the base augmentations (Pad+Crop+HFlip). This has two implications - firstly, this ensures that the complexity of these augmentations is always greater than or equal to the base augmentations. Secondly, since AutoAugment returns the unaugmented image as well, with a certain probability (0.22 for CIFAR-10 policy), the base augmentations form a subset of the complex augmentations. This trend is indeed reflected in the distribution of pair-wise feature-level similiarities (Cosine similarity between features obtained from an Inception-V3 network) between the following pairs: (Unaugmented, Pad+Crop+HFlip (PCHf)) and (Unaugmented, AutoAugment+PCHf) as shown in Fig.\ref{fig:aa_pc_hist}(a). 

When AutoAugment alone is applied, a large fraction of images have a very high cosine similarity, while others have a more spread out distribution as shown in Fig.\ref{fig:aa_pc_hist}(b). When Pad+Crop+HFlip is applied in series after AutoAugment, the distribution of these images shifts to the left, leading to an overlap in the two distributions. Even with such a large overlap the method works well because, the role of the "complex" batch-norm layer is to allow the learning of a function that minimizes empirical risk across a wide distribution of data. While the test distribution may be different from these augmentations, learning from diverse data is known to prevent overfitting and improve generalization. However, since the task of adversarial training is inherently hard, and the objective of minimizing loss on a wider distribution of data makes the task harder, we observe a drop in overall accuracy. The use of a separate batch-norm layer for "simple" augmentations allows the network to specialize on a select subset that is close to the distribution of test set images, and has a low variance. While the diversity of simple augmentations is low, it is sufficient to learn the batch-norm statistics and affine parameters which constitute 0.05\% of all parameters, while majority of the parameters are learned using both distributions, resulting in low overfitting.

\subsection{Ablations}

We show the impact of some of the important categories of augmentations individually in Table-\ref{table:augs}. 

\begin{table*}[h]
\caption{Impact of individual augmentations within AutoAugment \cite{cubuk2018autoaugment} on the proposed defense DAJAT. Performance on CIFAR-10 dataset with ResNet-18 architecture is reported. Robust evaluations are done against GAMA attack \cite{sriramanan2020gama}}
\setlength\tabcolsep{2pt}
\resizebox{1.0\linewidth}{!}{
\label{table:augs}
\begin{tabular}{l|c|c||c|c}
\toprule
\multicolumn{1}{l|}{}                       & \multicolumn{2}{c||}{\textbf{Augmentation}}             & \multicolumn{2}{c}{\textbf{Base + Augmentation + JS}} \\
\cmidrule{2-5}
\textbf{Augmentation}                      & \textbf{~~~~Clean Acc~~~~}        & \textbf{~~~~Robust Acc~~~~}       & \textbf{~~~~Clean Acc~~~~}        & \textbf{~~~~Robust Acc~~~~}       \\
\midrule

AutoAugment            & 82.54 & 48.11 & \textbf{84.94} & 51.23 \\
AutoAugment (without spatial augs)~~~~~~~ & \textbf{83.70}                      & 48.80                      & 84.94                     & 51.40    \\   
Brightness                                 & 82.11                     & 46.42                     & 84.56                     & 50.94                   \\
Sharpness                                  & 81.78                     & 49.85                     & 84.30                     & 50.54                     \\
Color-Balance                                     & 82.31                     & \textbf{49.87}                     & 84.39                     & \textbf{51.48}                     \\
Contrast                                   & 82.45                     & 46.54                     & 84.21                     & 50.77                     \\

\bottomrule

\end{tabular}}
\end{table*}

The robust accuracy improves when the spatial augmentations - shear, rotation and translation are not used for adversarial training. Amongst the augmentations that modify the low-level statistics of images, change in color balance gives maximum benefit. Further, it can be noted that although some augmentations such as change in brightness and contrast lead to a drop in robust accuracy when used directly, the use of the same augmentations in the proposed framework results in a significant boost in accuracy. Therefore there is no need to use carefully selected augmentations with the proposed framework.

\section{Split Batch-Normalization}
\label{app:splitbn}
\begin{table}[t]
\centering
\caption{\textbf{Split Batch-Normalization:} Impact of using common/ split running statistics and affine parameters. Using a combination of separate running statistics and affine parameters works the best.}
\vspace{0.1cm}
\setlength\tabcolsep{2pt}
\resizebox{1.0\linewidth}{!}{
\label{table:bn_additonal_one}
\begin{tabular}{l|c|c}
\toprule
\multicolumn{1}{l|}{\textbf{Method}}                                                & \multicolumn{1}{c|}{\textbf{Clean Accuracy}} & \textbf{Robust Accuracy} \\
\midrule
\multicolumn{1}{l|}{[E1] split running statistics + split affine parameters (Ours)} & \multicolumn{1}{c|}{\textbf{88.90}}                                    & \textbf{57.22}                                        \\ 

\multicolumn{1}{l|}{[E2] split running statistics + common affine parameters} & \multicolumn{1}{c|}{88.61}                                    & 56.91                                        \\ 

\multicolumn{1}{l|}{[E3] common running statistics + split affine parameters} & \multicolumn{1}{c|}{88.86}                                    & 57.01                                         \\ 

\multicolumn{1}{l|}{[E4] common running statistics + common affine parameters (single Batch-Norm)} & \multicolumn{1}{c|}{89.08}                                    & 53.86                              \\ 
\bottomrule
\end{tabular}}
\vspace{-0.1cm}
\end{table}

We perform three ablation experiments - first by training with separate running statistics and common affine parameters, second by training with common running statistics and separate affine parameters, and third without using split batch-norm at all, as shown in Table-\ref{table:bn_additonal_one}. Robust Accuracy is reported against the GAMA attack. As shown in the Table-\ref{table:bn_additonal_one}, E2 and E3 (having either split running statistics or split affine parameters) perform similar to the proposed approach, where separate running statistics and affine parameters are used. This shows that our method learns a different function mapping for both augmentations, and this can be realized by having different running statistics or different affine parameters or using a combination of both. We further note that the use of a single batch-norm layer for both augmentations (E4) degrades the results significantly.

\begin{figure}
\centering
        \includegraphics[width=0.6\linewidth]{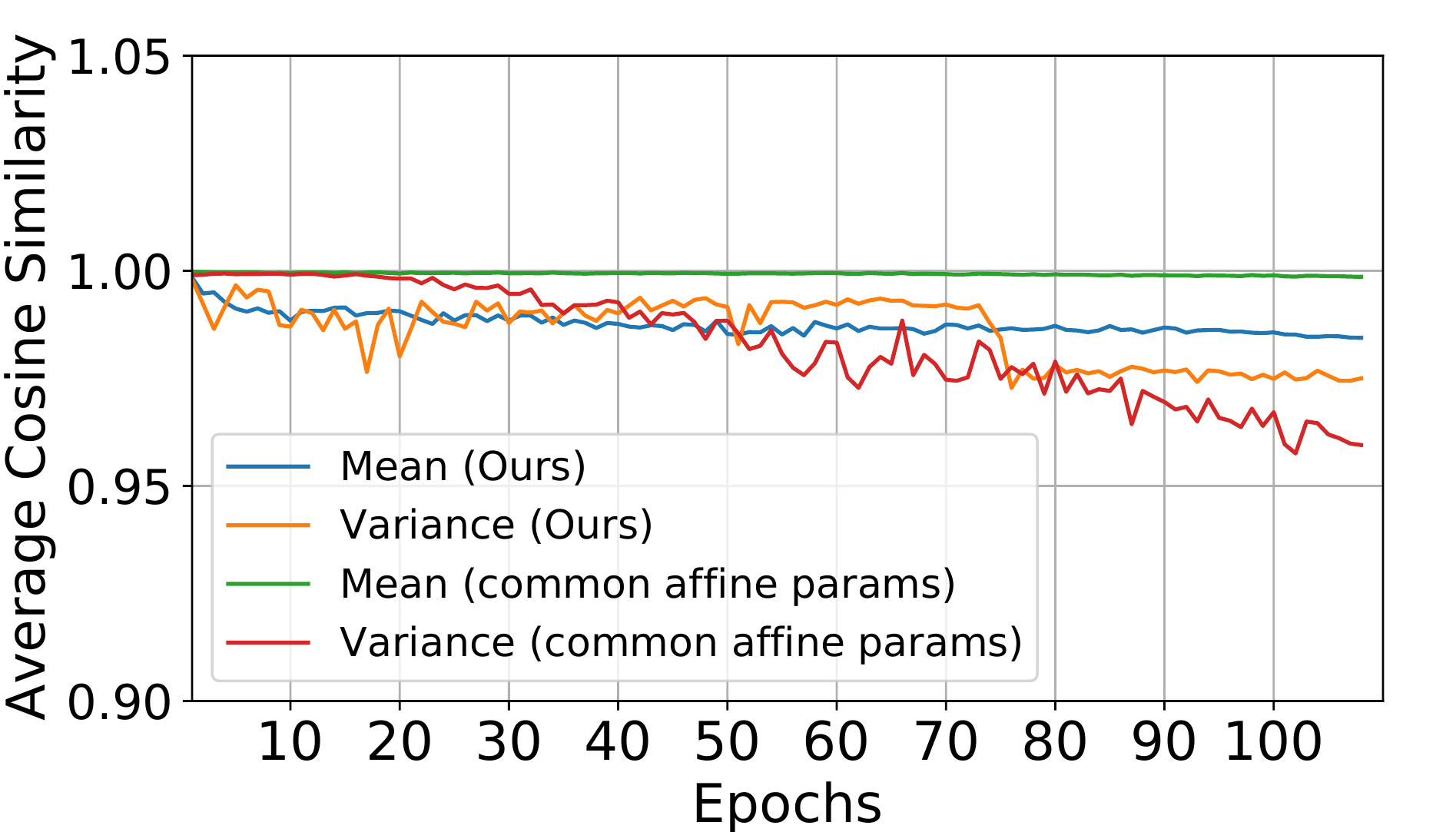}
        \caption{\textbf{Average Cosine Similarity (averaged over all the layers of WideResNet-34-10) of Running Mean and Variance} across training epochs. For the proposed approach DAJAT, and an ablation experiment that uses common batch-normalization affine parameters, the scale of cosine similarities are similar, indicating that small changes in running mean and variance can indeed have a large impact on model outputs.}
        \label{fig:bn_noaffine}
\end{figure}

We compare the average cosine similarity of the running statistics across training iterations for both our method, and the case where we have separate running statistics and common affine parameters in Fig.\ref{fig:bn_noaffine}. We note that the scale and trend of the average cosine similarity is similar in both cases. In the case where common affine parameters are used, the gains in results w.r.t. single batch-norm case can be attributed to the small drop in cosine similarity over training. This indicates that small changes in these running statistics can indeed lead to a large impact in the overall results (as shown in Table-\ref{table:bn_additonal_one}). We further verify this by noting the large difference in performance of the models when different batch-norm parameters are used for training, in Table-\ref{table:bn_additonal_two}. Robust Accuracy is reported against the GAMA attack.

\begin{table}[t]
\centering
\caption{\textbf{Impact of using different batch-normalization layers} (Pad+Crop+HFlip vs AutoAugment) \textbf{at inference time.}}
\vspace{0.1cm}
\setlength\tabcolsep{2pt}
\resizebox{1.0\linewidth}{!}{
\label{table:bn_additonal_two}
\begin{tabular}{l|c|c|c}
\toprule
 \multicolumn{1}{l|}{\textbf{Method~~~~~~~~~~~~~~~~~~~~}}                                           & \multicolumn{1}{l|}{\textbf{Batch-norm (Inference)}} & \textbf{Clean Acc}           & \textbf{Robust Acc}                    \\ 
 \midrule
 
 \multicolumn{1}{l|}{Split running statistics + split affine parameters (Ours)} & Pad+Crop+HFlip  & \multicolumn{1}{c|}{\textbf{88.90}}                             & \textbf{57.22}                                        \\ 

 \multicolumn{1}{l|}{Split running statistics + split affine parameters (Ours)} & AutoAugment  & \multicolumn{1}{c|}{76.17  }                                    & 44.45                                       \\                                        

 \multicolumn{1}{l|}{Split running statistics + common affine parameters} & Pad+Crop+HFlip  & \multicolumn{1}{c|}{88.61}                                    & 56.91                                     \\ 
 
 \multicolumn{1}{l|}{Split running statistics + common affine parameters} & AutoAugment  & \multicolumn{1}{c|}{78.69}                                    & 45.41                                       \\ 
\bottomrule
\end{tabular}}
\vspace{-0.1cm}
\end{table}

\begin{figure*}
\begin{minipage}{0.48\linewidth}
\centering
\includegraphics[width=\linewidth]{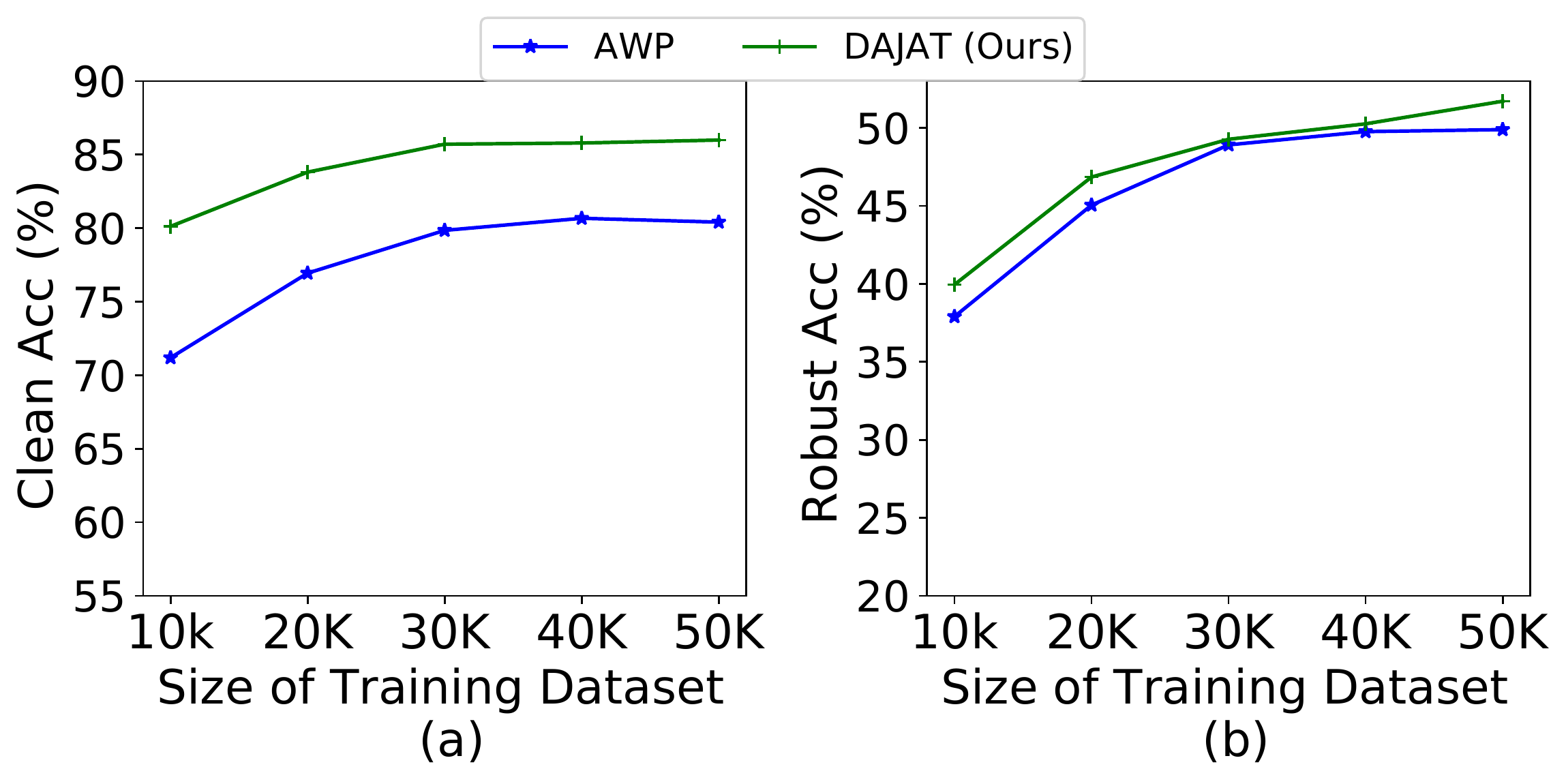}
\caption{\textbf{Improvements of DAJAT in the low data regime:} Performance of proposed approach DAJAT (Base, 2*AA) when compared to the TRADES-AWP \cite{wu2020adversarial} 2-step baseline on ResNet-18 architecture and CIFAR-10 dataset. The proposed approach achieves improvements across different sizes of training dataset, with higher gains in the low data regime.}
\label{fig:vary_data}
\end{minipage}
\hfill
\begin{minipage}{0.48\linewidth}
\centering
\includegraphics[width=\linewidth]{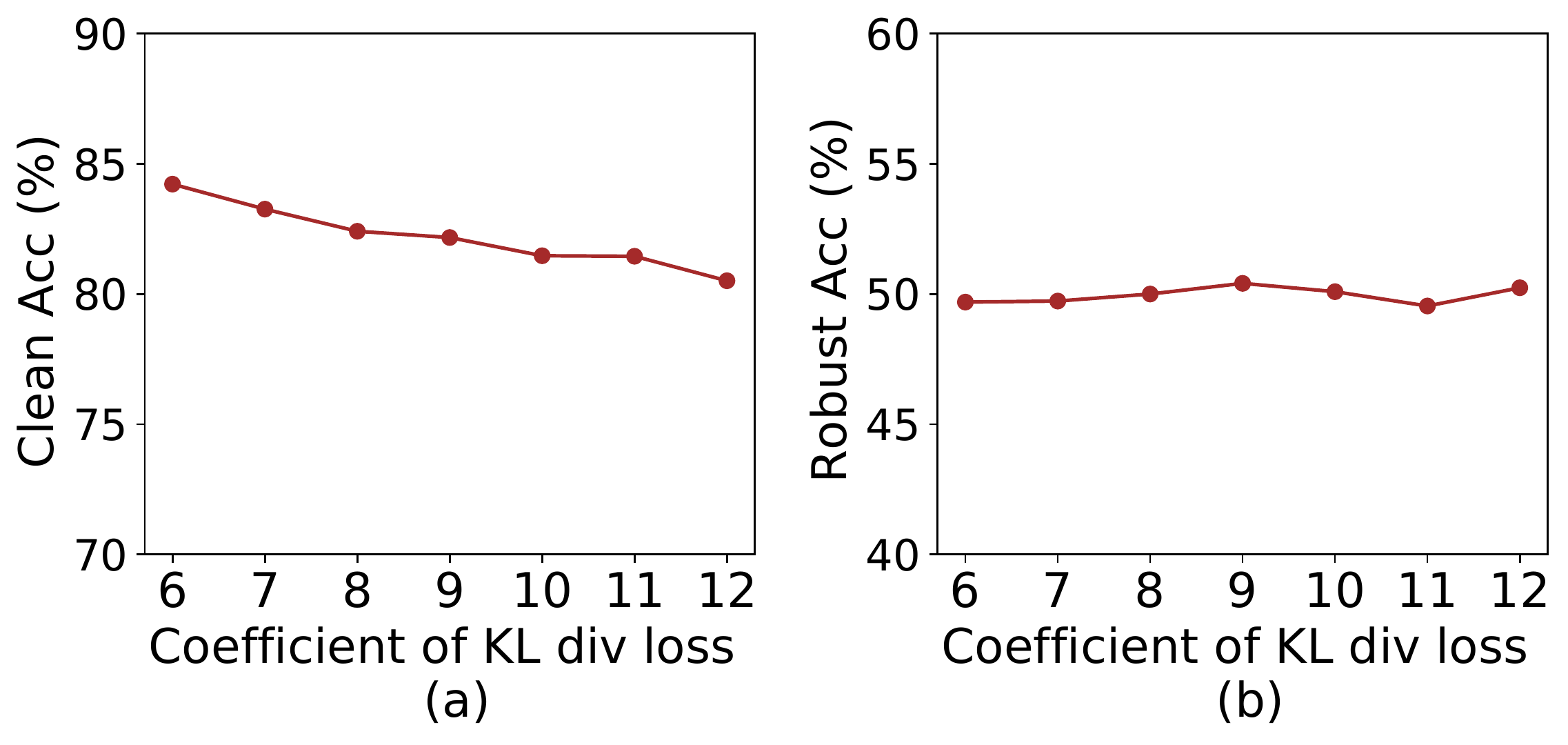}
\caption{\textbf{Performance of ACAT across variation in $\beta$: }Performance of the proposed defense ACAT against variation in the coefficient of KL divergence loss between the clean and adversarial samples $\beta$. Higher $\beta$ leads to improved robust accuracy (against GAMA \cite{sriramanan2020gama} attack) at the cost of clean accuracy. ResNet-18 architecture and CIFAR-10 dataset are used. }
\label{fig:acat_sens}
\end{minipage}
\vspace{-0.6cm}
\end{figure*}

\section{Improvements of DAJAT in the low data regime}
We compare the performance of the proposed DAJAT defense (using Base, 2*AA) with TRADES-AWP \cite{wu2020adversarial} 2-step baseline across different sizes of the CIFAR-10 training dataset on ResNet-18 architecture in Fig.\ref{fig:vary_data}. We consider class balanced dataset for each case. We note that across different settings, the proposed approach achieves 5-9\% gains in clean accuracy and an average gain of 1.3\% in robust accuracy. The gain in clean accuracy in the low data regime is significantly high (8.93\%) highlighting the need for augmentations in improving Adversarial Training performance, specifically in real-world settings where the training data is low. 

\section{Details on ACAT}

\subsection{Motivation: Instability of 2-step Adversarial Training}
\label{appsubsec:motivacat}
Ascending perturbation radius helps in mitigating gradient masking when lesser steps are used for attack generation. We present results of ACAT when compared to Fixed constraint AT (TRADES-AWP with 2 attack steps) on CIFAR-10 dataset using $\varepsilon=8/255$ with ResNet-18 and WideResNet-34-10 model architectures in Table-\ref{table:vareps_abl_two}. Robust Accuracy is reported against the GAMA attack. Best accuracy is computed using PGD-20 attack, which is not very reliable. Hence, in some cases, best epoch may have a slightly lower accuracy when compared to the last epoch. On ResNet-18, we observe that the difference between last and best epochs for both methods is very low. However, on WideResNet-34-10, we observe the phenomenon of gradient masking in Fixed Constraint AT, with robust accuracy dropping by around 10\% towards the end of training. We note that the difference between last and best epochs is very low for ACAT even on WideResNet-34-10. The main motivation of using an ascending perturbation radius is to stabilize training and prevent the onset of gradient masking.
\begin{table}[t]
\centering
\caption{Fixed vs ascending $\varepsilon$ Adversarial training on ResNet-18 and WRN-34-10 models with maximum $\varepsilon = 8/255$ on CIFAR-10 dataset with 2 step attack generation during training. Impact of gradient masking is higher for larger capacity models. This is effectively mitigated with the use of Ascending $\varepsilon$ Adversarial Training}
\vspace{0.1cm}
\setlength\tabcolsep{2pt}
\resizebox{1.0\linewidth}{!}{
\label{table:vareps_abl_two}
\begin{tabular}{l|c|c|c|c|c|c|c}
\toprule
\multicolumn{1}{l|}{\textbf{Architecture}}             & \multicolumn{1}{c|}{\textbf{Method}}    & \textbf{Clean(last)}     & \textbf{Robust(last)} & \textbf{Clean(best)} & \textbf{Robust(best)} & \textbf{Clean(last - best)} & \textbf{Robust(last - best)} \\
\midrule
\multicolumn{1}{l|}{ResNet-18} & \multicolumn{1}{c|}{Fixed $\varepsilon$ AT} & 80.63 & 49.63                      & 80.82                     & 49.61                      & -0.19                        & 0.02                              \\
ResNet-18                                               & Ascending $\varepsilon$ AT                                          & \textbf{82.41}                & \textbf{50.00}             & \textbf{82.57}            & \textbf{49.91}             & -0.16                        & 0.09                              \\
\midrule
WRN-34-10                                        & Fixed $\varepsilon$ AT                                              & 86.69                         & 44.87                      & \textbf{86.83}            & 54.76                      & -0.14                        & -9.89                             \\
WRN-34-10                                        & Ascending $\varepsilon$ AT                                          & \textbf{86.71}                & \textbf{55.58}             & 86.30                     & \textbf{55.46}             & 0.41                         & 0.12  \\
\bottomrule
\end{tabular}}
\end{table}

\subsection{ACAT training algorithm}
\label{sec:acat_algo}
The algorithm for the proposed ACAT defense is presented in Algorithm-\ref{alg:ACAT}. We consider an $\ell_\infty$ threat model of perturbation radius $\varepsilon$. The perturbation bound for attack generation $\varepsilon_{\mathrm{asc}}$ is linearly increased from $0$ to $\varepsilon$ during training (L3). The learning rate follows a cosine schedule across the training epochs as shown in L4. The attack generation (L7-L12) is done for $2$ iterations and follows the TRADES \cite{zhang2019theoretically} settings. Initially Gaussian noise of magnitude $0.001$ is added to every pixel. The KL divergence loss between the clean and perturbed images is maximized using the perturbation step size $\varepsilon_{\mathrm{asc}}$. Further, the perturbation is clipped to remain within the threat model in every iteration. 

As shown in L14, the TRADES-AWP \cite{wu2020adversarial} loss is used for  adversarial training. The loss $\mathcal{L}_{\mathrm{TR}}(\theta+\hat{\theta})$ is maximized with respect to $\hat{\theta}$ to find the perturbation $\tilde{\theta}$ within the constraint set $\mathcal{M}(\theta)$ (L15). Further, the model weights $\theta$ are update using gradients at $f_{\theta+\tilde{\theta}}$ (L16). The defense ACAT does not use any additional training hyperparameters when compared to the TRADES-AWP defense. We vary the hyperparameter $\beta$ to obtain optimal results. We show  the trend of model performance across variation in $\beta$ in Section-\ref{sec:hyp_sens}.

\subsection{Analysis of the proposed ACAT defense} 
\label{subsec:acat_analysis_supple}
\vspace{0.2cm}
\noindent \textbf{Conjecture-2:} We hypothesize that given a Neural Network $f_{\theta_\varepsilon}$ that minimizes the TRADES loss \cite{zhang2019theoretically} within a maximum perturbation radius of $\varepsilon$, and has sufficient smoothness in weight space within an $\ell_2$ radius $\psi$ around $\theta_\varepsilon$, there exist $\varepsilon’ > \varepsilon$ and a model $f_{\theta_{\varepsilon'}}$ where $||\theta_{\varepsilon'} - \theta_\varepsilon|| \leq \psi$, such that $f_{\theta_{\varepsilon'}}$  has a lower TRADES loss within $\varepsilon'$ when compared to $f_{\theta_{\varepsilon}}$.

\vspace{0.2cm}
\noindent \textbf{Justification:} We first consider Adversarial Training within a perturbation bound $\varepsilon$ using the TRADES-AWP loss shown below.  
\abovedisplayskip=-5pt

\begin{equation}
    \mathcal{L}_{\mathrm{AWP},\theta,\varepsilon} = \max\limits_{\hat{\theta} \in \mathcal{M}(\theta)} \frac{1}{N} \sum\limits_{i=1}^N \mathcal{L}_\mathrm{CE}(f_{\theta+\hat{\theta}}(x_i),y_i) + \beta \cdot \max\limits_{\tilde{x}_i \in \mathcal{A_{\varepsilon}}(x_i)} \mathrm{KL}(f_{\theta+\hat{\theta}}(x_i)||f_{\theta+\hat{\theta}}(\tilde{x}_i))~~~~~~~~~~~~~~~~~~~~~~~~~~~~~~~~~~~~~~~
\end{equation}

\begin{equation}
\label{eq:argmin}
    {\theta_\varepsilon} = \argmin \limits_\theta \mathcal{L}_{\mathrm{AWP},\theta,\varepsilon}~~~~~~~~~~~~~~~~~~~~~~~~~~~~~~~~~~~~~~~~~~~~~~~~~~~~~~~~~~~~~~~~~~~~~~~~~~~~~~~~~~~~~~~~~~~~~~~~~~~~~~~~~~~~~~~~~~~~~~~~~~~~~~~~~~~~~~~~~~~~~~
\end{equation}

The model $f_{{\theta_\varepsilon}}$ which is the minimizer of the above loss, achieves an optimal trade-off between the cross-entropy loss on clean samples (clean loss, $\mathcal{L}_\mathrm{clean}$) and weighted KL divergence between clean and adversarial images \cite{zhang2019theoretically} (adversarial loss, $\mathcal{L}_\mathrm{adv}$).

\abovedisplayskip=-10pt

\begin{equation}
    \mathcal{L}_{\mathrm{clean},\theta} = \sum\limits_{i=1}^N \mathcal{L}_\mathrm{CE}(f_{{\theta}}(x_i),y_i),~~~\mathcal{L}_{\mathrm{adv},\theta,\varepsilon} = \beta \cdot  \sum\limits_{i=1}^N  \mathrm{KL}(f_{{\theta}}(x_i)||f_{{\theta}}(\tilde{x}_i)),~~ \tilde{x} \in \mathcal{A}_\varepsilon(x)~~~~~~~~~~~~~~~~~~~~~~~~~~~~~~~~~
\end{equation}
Since the loss attains a minima at $\theta_\varepsilon$, a direction of further reduction in both clean and adversarial losses does not exist (Eq.\ref{eq:argmin}). Gradient descent on the adversarial loss at $\theta_\varepsilon$ to parameter $\theta'$ results in a reduction in adversarial loss at the cost of a higher clean loss. 

\abovedisplayskip=-5pt

\begin{equation}
\label{eq:grad_desc}
    \theta' = \theta_\varepsilon - \eta \cdot \nabla \mathcal{L}_{\mathrm{adv},\theta_\varepsilon,\varepsilon}~~~~~~~~~~~~~~~~~~~~~~~~~~~~~~~~~~~~~~~~~~~~~~~~~~~~~~~~~~~~~~~~~~~~~~~~~~~~~~~~~~~~~~~~~~~~~~~~~~~~~~~~~~~~~~~~~~~~~~~~~~~~~~~
\end{equation}
\abovedisplayskip=-5pt
\begin{equation}
    \mathcal{L}_{\mathrm{adv},\theta',\varepsilon} < \mathcal{L}_{\mathrm{adv},\theta_\varepsilon,\varepsilon},~~~\mathcal{L}_{\mathrm{clean},\theta'} > \mathcal{L}_{\mathrm{clean},\theta_\varepsilon} ~~~~~~~~~~~~~~~~~~~~~~~~~~~~~~~~~~~~~~~~~~~~~~~~~~~~~~~~~~~~~~~~~~~~~~~~~~~~~~~~~~~~~~~~~~~~~~~~~~~~~
\end{equation}

We assume that $\mathcal{L}_{\mathrm{clean},\theta_\varepsilon}$ increases by a rate $\gamma_{\mathrm{CE},\theta_\varepsilon}$ and $\mathcal{L}_{\mathrm{adv},\theta_\varepsilon,\varepsilon}$ decreases by a rate $\gamma_{\mathrm{KL_\varepsilon},\theta_\varepsilon}$ in the direction of gradient descent of the adversarial loss ( $-\nabla \mathcal{L}_{\mathrm{adv},\theta_\varepsilon,\varepsilon})$.
Since $\theta_\varepsilon$ is a minimizer of loss, 

\abovedisplayskip=10pt
\belowdisplayskip=5pt
\begin{equation}
\label{eq:alpha}
    \mathcal{L}_{\mathrm{AWP},\theta',\varepsilon} > \mathcal{L}_{\mathrm{AWP},\theta_\varepsilon,\varepsilon} \implies \gamma_{\mathrm{CE},\theta_\varepsilon} = \gamma_{\mathrm{KL_\varepsilon},\theta_\varepsilon} + \alpha, ~~\alpha > 0 ~~~~~~~~~~~~~~~~~~~~~~~~~~~~~~~~~~~~~~~~~~~~~~~~~~~~~~~~~~~~~~~
\end{equation}

Since the Cross-entropy loss is minimized at $x$, it attains a minima at $x$ within the pixel neighborhood. 
\begin{equation}
    \mathcal{L}_{\mathrm{CE}}(f_{\theta_\varepsilon}(x),y)) < \mathcal{L}_{\mathrm{CE}}(f_{\theta_\varepsilon}(x+\epsilon),y)~~~~~~~~~~~~~~~~~~~~~~~~~~~~~~~~~~~~~~~~~~~~~~~~~~~~~~~~~~~~~~~~~~~~~~~~~~~~~~~~~~~~~~~~~~~~~~~~~~~~~~~~~~~~~~~~~~~
\end{equation}

Since the model $f_\theta$ is adversarially robust within $\varepsilon$, the loss surface is locally smooth within the open interval $(-\varepsilon,\varepsilon)$. The adversarial loss, which measures the KL divergence of the adversarial image w.r.t. the clean image is least at $x$ and increases monotonically in the $\varepsilon$-neighborhood for adversarially robust models \cite{athalye2018obfuscated}. Therefore, adversarial loss at $\varepsilon'$ = $\varepsilon$ + $\delta$ for a small enough $\delta$ would be marginally higher than the loss at $\varepsilon$. 
\begin{equation}
     \mathcal{L}_{\mathrm{adv},\theta_\varepsilon,\varepsilon'} > \mathcal{L}_{\mathrm{adv},\theta_\varepsilon,\varepsilon} ~~~~~~~~~~~~~~~~~~~~~~~~~~~~~~~~~~~~~~~~~~~~~~~~~~~~~~~~~~~~~~~~~~~~~~~~~~~~~~~~~~~~~~~~~~~~~~~~~~~~~~~~~~~~~~~~~~~~~~~~~~~~~~~~~~~~~~~~~~~~
\end{equation}

Based on the local smoothness in weight space as well, the same property holds at any $\theta'$ in the gradient descent direction of $\mathcal{L}_{\mathrm{adv},\theta_\varepsilon,\varepsilon}$ (satisfying Eq.\ref{eq:grad_desc}) such that $||\theta’ - \theta_\varepsilon|| \leq \psi$, as shown below:
\begin{equation}
   \mathcal{L}_\mathrm{adv,\theta_\varepsilon,\varepsilon'} = \mathcal{L}_\mathrm{adv,\theta_\varepsilon,\varepsilon} + \delta_{\theta_\varepsilon} ~~~~~~~~~~~~~~~~~~~~~~~~~~~~~~~~~~~~~~~~~~~~~~~~~~~~~~~~~~~~~~~~~~~~~~~~~~~~~~~~~~~~~~~~~~~~~~~~~~~~~~~~~~~~~~~~~~~~~~~
\end{equation}
\abovedisplayskip=-10pt
\belowdisplayskip=10pt
\begin{equation}
    \mathcal{L}_\mathrm{adv,\theta',\varepsilon'} = \mathcal{L}_\mathrm{adv,\theta',\varepsilon} + \delta_{\theta'}~~~~~~~~~~~~~~~~~~~~~~~~~~~~~~~~~~~~~~~~~~~~~~~~~~~~~~~~~~~~~~~~~~~~~~~~~~~~~~~~~~~~~~~~~~~~~~~~~~~~~~~~~~~~~~~~~~~~~~~~~~~~~~~~
\end{equation}
Since the adversarial loss at $\theta'$ is lower than the same at $\theta_\varepsilon$, the loss surface at $\theta'$ has a lower local Lipschitz constant when compared to $\theta_\varepsilon$ leading to the following: 
\abovedisplayskip=8pt
\belowdisplayskip=8pt
\begin{equation}
    \delta_{\theta'} < \delta_{\theta_\varepsilon} \implies \mathcal{L}_\mathrm{adv,\theta',\varepsilon'} - \mathcal{L}_\mathrm{adv,\theta',\varepsilon} < \mathcal{L}_\mathrm{adv,\theta_\varepsilon,\varepsilon'} - \mathcal{L}_\mathrm{adv,\theta_\varepsilon,\varepsilon} ~~~~~~~~~~~~~~~~~~~~~~~~~~~~~~~~~~~~~~~~~~~~~~~~~~~~~~~~~~~~~~~
\end{equation}
Rearranging terms,
\begin{equation}
    \mathcal{L}_\mathrm{adv,\theta_\varepsilon,\varepsilon'} - \mathcal{L}_\mathrm{adv,\theta',\varepsilon'} > \mathcal{L}_\mathrm{adv,\theta_\varepsilon,\varepsilon} - \mathcal{L}_\mathrm{adv,\theta',\varepsilon} ~~~~~~~~~~~~~~~~~~~~~~~~~~~~~~~~~~~~~~~~~~~~~~~~~~~~~~~~~~~~~~~~~~~~~~~~~~~~~~~~~~~~~~~~~~~~~~~~~~~~~~~~~~~~~~~~~~~~~~~
\end{equation}
For small enough gradient descent step size $\eta$, the above can be related to the rate of reduction in the adversarial loss as shown below:
\begin{equation}
\label{eq:alpha_dash}
    \gamma_{\mathrm{KL}_{\varepsilon'},\theta_\varepsilon} = \gamma_{\mathrm{KL}_\varepsilon,\theta_\varepsilon} + \alpha', ~~~ \alpha' > 0   ~~~~~~~~~~~~~~~~~~~~~~~~~~~~~~~~~~~~~~~~~~~~~~~~~~~~~~~~~~~~~~~~~~~~~~~~~~~~~~~~~~~~~~~~~~~~~~~~~~~~~~~~~~~~~~~~~~~~~~~~~~~~~~~~~~~~~~~~~~~~~~
\end{equation}
Since the Cross-entropy loss does not depend on $\varepsilon$, 

and Eq.\ref{eq:alpha_dash},
\begin{equation}
    \gamma_{\mathrm{CE},\theta_\varepsilon} - \gamma_{\mathrm{KL}_{\varepsilon'},\theta_\varepsilon} = \alpha - \alpha' ~~~~~~~~~~~~~~~~~~~~~~~~~~~~~~~~~~~~~~~~~~~~~~~~~~~~~~~~~~~~~~~~~~~~~~~~~~~~~~~~~~~~~~~~~~~~~~~~~~~~~~~~~~~~~~~~~~~~~~~~~~
\end{equation}
Therefore, the rate of increase in $\mathcal{L}_{\mathrm{AWP},\theta_\varepsilon,\varepsilon'}$, or $\alpha-\alpha'$, is less than the rate of increase in  $\mathcal{L}_{\mathrm{AWP},\theta_\varepsilon,\varepsilon}$, or $\alpha$. For small enough $\alpha$, $\alpha' > \alpha$, so that the overall loss $\mathcal{L}_{\mathrm{AWP},\theta_\varepsilon,\varepsilon'}$ decreases. 

Based on this, for small enough $\eta$, $\exists~ \theta_{\varepsilon'}~ s.t.~||\theta_{\varepsilon'} - \theta_\varepsilon|| \leq \psi$ and $\mathcal{L}_\mathrm{AWP,\theta_{\varepsilon'},\varepsilon'} < \mathcal{L}_\mathrm{AWP,\theta_\varepsilon,\varepsilon'}$. Hence, it is possible to move to $\theta_{\varepsilon'}$ which has a lower overall loss than the loss at $\theta_\varepsilon$. $\hfill\blacksquare$

\begin{algorithm}[tb]
   \caption{Ascending Constraint Adversarial Training (ACAT)}
   \label{alg:ACAT}
   
\begin{algorithmic}[1]
   \STATE {\bfseries Input:} Network $f_\theta$, Training Dataset $\mathcal{D} =\{(x_i,y_i) \}$, Adversarial Threat model: $\ell_\infty$ bound of radius $\varepsilon$, number of epochs E, Maximum Learning Rate $\mathrm{LR}_{max}$, $M$ training mini-batches of size $n$, Cross-entropy loss $\ell_{CE}$, Weight perturbation constraint $\mathcal{M}(\theta)$, coefficient of KL divergence term $\beta$

\FOR{$epoch=1$ {\bfseries to} $E$}
    \STATE $\varepsilon_{\mathrm{asc}} = epoch \cdot \varepsilon/ E $
    \STATE $\mathrm{LR} = 0.5  \cdot \mathrm{LR}_{max} \cdot (1 + cosine((epoch -1)/E\cdot \pi))$
    \FOR{$iter=1$ {\bfseries to} $M$}
        \FOR{$i=1$ {\bfseries to} $n$ (in parallel)} 
        \FOR{$steps=1$ {\bfseries to} $2$}
        \STATE $\delta = 0.001 \cdot \mathcal{N}(0,1)$
        \STATE $\delta = \delta + \varepsilon_{\mathrm{asc}} \cdot \mathrm{sign}\left(\nabla_{\delta} \mathrm{KL}(f_{{\theta}}(x_i)||f_{{\theta}}(x_i+\delta)) \right)$
        \STATE $\delta = Clamp~(\delta,-\varepsilon_{\mathrm{asc}},\varepsilon_{\mathrm{asc}})$
        \STATE $\widetilde{x}_i = Clamp~(x_i + \delta,0,1)$
        \ENDFOR
        \ENDFOR
         \STATE $\mathcal{L}_{\mathrm{TR}}(\theta) =  \frac{1}{n} \sum\limits_{i=1}^n \mathcal{L}_\mathrm{CE}(f_{{\theta}}(x_i),y_i) + \beta \cdot  \mathrm{KL}(f_{{\theta}}(x_i)||f_{{\theta}}(\tilde{x}_i))$
         \STATE $\tilde{\theta} = \argmax\limits_{{\hat{\theta}} \in \mathcal{M}(\theta)} \mathcal{L}_{\mathrm{TR}}(\theta+\hat{\theta})$
         \STATE $\theta = \theta -  \mathrm{LR} \cdot \nabla_{\theta} ( \mathcal{L}_{\mathrm{TR}}(\theta+\tilde{\theta}))$

        \ENDFOR

\ENDFOR
\end{algorithmic}
\end{algorithm}
\vspace{-0.3cm}

\vspace{0.1cm}
\subsection{Integrating ACAT with other efficient training methods}

\label{subsec:nuat_gat_supple}
The proposed ACAT defense uses the KL divergence loss between clean and adversarial images, similar to the TRADES adversarial training algorithm \cite{zhang2019theoretically}. We present results by integrating the proposed ACAT defense with losses from existing efficient adversarial training algorithms \cite{sriramanan2020gama,sriramanan2021nuat} in Table-\ref{table:acat_gat_nuat}. We obtain a significant boost in performance over the respective baselines, when we use ACAT with GAT \cite{sriramanan2020gama} and TRADES \cite{zhang2019theoretically} losses, and a marginal boost when integrated with the NuAT defense \cite{sriramanan2021nuat}. The adversarial weight perturbation step in the proposed defense results in an increase in computational time when compared to the respective baselines. We choose the KL divergence based loss for both proposed defenses ACAT and DAJAT since it results in an optimal trade-off between performance and training time. 
\begin{table}[H]
\centering
\caption{\textbf{Integrating ACAT with different loss formulations} on the CIFAR-10 dataset with WideResNet-34-10 architecture. Robust accuracy is reported against the GAMA attack \cite{sriramanan2020gama}.}
\vspace{0.1cm}
\setlength\tabcolsep{2pt}
\label{table:acat_gat_nuat}
\resizebox{0.85\linewidth}{!}{
\begin{tabular}{l|c|c|c|c}
\toprule

\multicolumn{1}{l|}{\textbf{Method}} &\textbf{ \#{ Attack Steps}} & \textbf{Clean Acc} & \textbf{Robust Acc} & \textbf{Time per epoch (seconds)}  \\
\midrule

TRADES-AWP          & 2        & 85.49          & 41.62 & 412 \\
ACAT (with TRADES loss)     & 2        & 86.71          & 55.58  & 412 \\
NuAT2-WA             & 2        & 86.32          & 55.08 & 334 \\
ACAT (with NuAT loss)      & 2        & 86.19          & \textbf{55.91} & 530 \\
GAT2-WA              & 2        & 87.36          & 50.24 & 267 \\
ACAT (with GAT loss)       & 2        & \textbf{87.79}          & 54.70 & 396 \\
\bottomrule
\end{tabular}
}
\vspace{-0.4cm}
\end{table}
\begin{table}[]
\centering
\caption{Impact of augmentations: Performance ($\%$) of ACAT models on Base augmentations and AutoAugment (Auto). Clean and robust accuracy against GAMA attack \cite{sriramanan2020gama} are reported. The use of AutoAugment results in $\sim 1.5$ - $2\%$ drop in robust accuracy. The use of Base Augmentations alone (Pad+Crop+HFlip) gives the best overall performance on the unaugmented test set.}
\vspace{0.1cm}
\resizebox{1.0\linewidth}{!}{
\label{table:acat_additional}
\begin{tabular}{l|c|c|c|c|c}
\toprule
\multicolumn{1}{l|}{\textbf{Architecture}}           & \multicolumn{1}{c|}{\textbf{Train set}}      & \multicolumn{1}{l|}{\textbf{No-Aug (Clean)}} & \multicolumn{1}{l|}{\textbf{No-Aug (Robust)}} & \multicolumn{1}{l|}{\textbf{AutoAug (Clean)}} & \multicolumn{1}{l}{\textbf{AutoAug (Robust)}} \\ 
\midrule
                      & \multicolumn{1}{c|}{No-Augmentation}         & 73.50                                       & 43.64                                        & 44.98                                        & 18.50                                         \\ 
 
 & Base (Pad+Crop+HFlip)                        & 82.41                                       & 50.00                                        & 63.79                                        & 37.07                                         \\
ResNet-18                       & AutoAugment                                  & \textbf{82.54}                                       & 48.11                                        & \textbf{76.40}                                        & \textbf{43.22}                                         \\
                   & Base (50\% batch) + AutoAugment (50\% batch) & 81.15                                       & \textbf{50.01}                                        & 70.89                                        & 40.93                                         \\
\midrule
                                     & No-Augmentation                              & 80.34                                       & 47.98                                        & 54.66                                        & 26.44                                         \\
 
                                    & Base (Pad+Crop+HFlip)                        & 86.71                                       & \textbf{55.58}                                        & 68.24                                        & 40.83                                         \\
WideResNet-34-10                                      & AutoAugment                                  & \textbf{86.80}                                       & 53.99                                        & \textbf{82.64}                                        & \textbf{48.98}                                         \\
                                     & Base (50\% batch) + AutoAugment (50\% batch) & 86.52                                       & 54.15                                        & 75.90                                        & 45.56                                            \\
\bottomrule
\end{tabular}}
\vspace{-0.1cm}
\end{table}
\subsection{Analysis of the effect of augmentations on ACAT} We analyze how using different combinations of augmentations affect ACAT in Table-\ref{table:acat_additional}. From these results we can say that using hard augmentations like AutoAugment and a mix of base and AutoAugment each on $50\%$ on the batch leads to a degradation in the performance of the model as compared to using simple augmentations like Pad$+$Crop$+$Horizontal Flip.

\section{Details on Experiments and Results}

\subsection{Details on Datasets}
\label{sec:dataset_details}
We perform evaluations of the proposed defenses ACAT and DAJAT on the CIFAR-10, CIFAR-100 \cite{krizhevsky2009learning} and ImageNette-10 \cite{howard2020fastai,imagenet_cvpr09} datasets, comprising of 10, 100 and 10 classes respectively. The resolution of images in the CIFAR-10 and CIFAR-100 datasets is 32x32, while it is 128x128 in ImageNette dataset. We use ResNet-18 \cite{RN18} and WideResNet-34-10 \cite{zagoruyko2016wide} architectures for the CIFAR-10 and CIFAR-100 datasets and ResNet-18 architecture for ImageNette. While CIFAR-10 is the most popular dataset used for benchmarking adversarial defenses, CIFAR-100 has a larger number of classes with one-tenth the number of images in each class, making it a more challenging setting. ImageNette dataset is used to show the performance of the proposed method on higher resolution images. We consider an $\ell_\infty$ threat model with $\varepsilon=8/255$ for the primary evaluations across all datasets. We use TRADES-AWP \cite{wu2020adversarial} \footnote{https://github.com/csdongxian/AWP} as the base code for most of our analysis. For analysis in Table-\ref{table:dajat_with_others} we use HAT \cite{rade2022reducing} \footnote{https://github.com/imrahulr/hat} and OAAT \cite{addepalli2021oaat}\footnote{https://github.com/val-iisc/OAAT} as the base codes. For cutmix analysis in Table-\ref{table:cutmix} we use \cite{rade2021pytorch}  \footnote{https://github.com/imrahulr/adversarial\_robustness\_pytorch} as the base code. The license for each of these codes are available on their respective github repositories. Since the datasets used in this work are public, commonly used for research purposes and do not contain any objectionable content, we find no need to take any consent from the authors of these datasets.

\begin{figure}
\centering
\includegraphics[width=\linewidth]{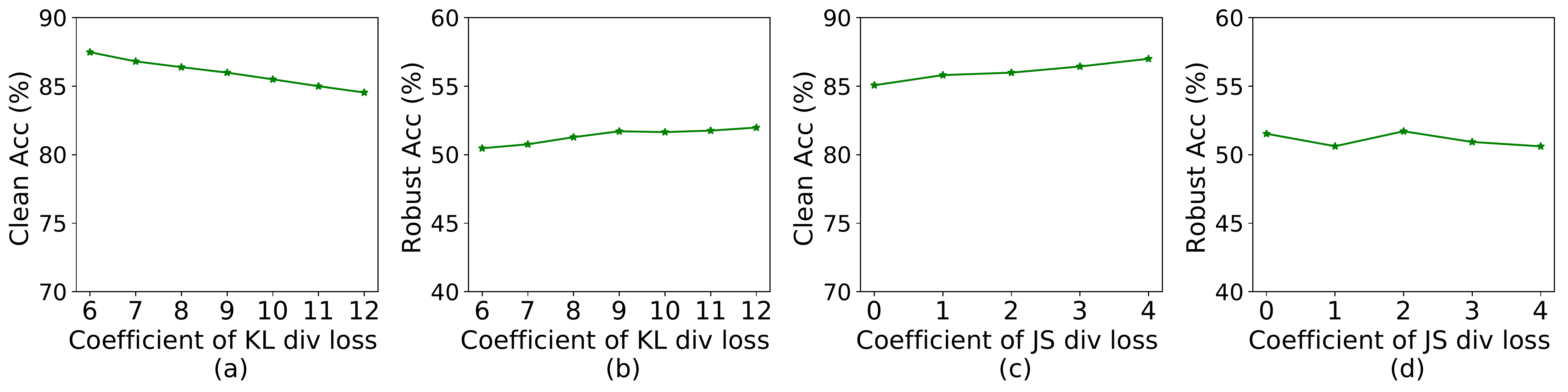}
\caption{\textbf{Performance of DAJAT across variation in hyperparameters:} Performance of the proposed defense DAJAT (Base, 2*AA) on ResNet-18 architecture and CIFAR-10 dataset across variation in (a,b) coefficient of KL divergence loss between softmax outputs of clean and adversarial images. (c,d) coefficient of the JS divergence loss between the representations of different augmentations. Robust accuracy against GAMA \cite{sriramanan2020gama} attack is shown.}
\label{fig:dajat_sens}
\end{figure}

\subsection{Details on Training Settings}
\label{sec:hyp_sens}

The training for the baseline methods and the proposed approach is done for 110 epochs as is common in literature \cite{pang2020bag}. We use SGD optimizer along with cosine learning rate schedule for the proposed method with a momentum of 0.9. We fix the maximum learning rate to 0.2 and weight decay to 5e-4 for the proposed approach across all datasets and model architectures. For the baselines, we use the training settings from the official codes released by the authors. We performed all the experiments on two NVidia V100 GPUs, two RTX-2080 GPUs and one RTX-3090 GPU. We use a validation split of 1000 images from the train dataset both for CIFAR-10 and CIFAR-100 images,
except for the results reported in Table-\ref{table:200_epochs}, where we use the full dataset to compare against the state-of-the-art defenses on the RobustBench leaderboard \cite{croce2021robustbench}.

For the proposed approaches ACAT and DAJAT (Base, 2*AA), we vary the value of $\beta$ to achieve an optimal trade-off between the clean and adversarial accuracy. As shown in Fig.\ref{fig:acat_sens} and Fig.\ref{fig:dajat_sens}(a,b), as we increase $\beta$, robust accuracy improves and clean accuracy degrades initially, with a saturating trend in robust accuracy at higher values of $\beta$. For the ResNet-18 model and CIFAR-10 dataset, the optimal value of $\beta$ is 8 and 9 for ACAT and DAJAT respectively. 

We further fix the value of $\beta$ to the optimal setting of 9 and vary the coefficient of the JS divergence term in Fig.\ref{fig:dajat_sens}(c,d). This term leads to a boost in the clean accuracy at the cost of a slight degradation in the robust accuracy. The optimal setting of the coefficient of JS divergence term is 2 in the given setting (Base, 2*AA) and ranges from 1 to 3 across all settings, datasets and model architectures. 

We consider the Base, 2*AA as the main setting of DAJAT since its computational complexity matches with that of TRADES-AWP. However, we show the result of Base, 3*AA as well to highlight that the performance improves with a further increase in diversity. In this case, since the weight of the Base augmentations is considerably low in the overall loss (L16 in Algorithm-\ref{alg:DAJAT}), we give a weight of 1/3 to the TRADES loss on base augmentations and 2/3 to the TRADES loss on the AutoAugment based images. This mimics the setting of Base, 2*AA in terms of loss weighting, while introducing additional diversity due to the presence of a larger number of complex augmentations, yielding a small boost in performance. 

We share the optimal set of hyperparameters across all datasets and model architectures in the ReadMe file that is submitted along with our Code in the Supplementary submission. 

\subsection{Performance on Larger Capacity Models}
\label{appsubsec:largecap}
\begin{table}[t]
\centering
\caption{Performance gains obtained using DAJAT on CIFAR-100 using \textbf{larger capacity models}}
\vspace{0.1cm}
\setlength\tabcolsep{2pt}
\resizebox{1.0\linewidth}{!}{
\label{table:SOTA_WRN3420}
\begin{tabular}{l|c|c|c|c}
\toprule
\multicolumn{1}{c|}{\textbf{Training Algorithm}} & \multicolumn{1}{l|}{\textbf{Architecture}}               & \multicolumn{1}{l|}{\textbf{Clean Accuracy}} & \multicolumn{1}{l|}{\textbf{Robust Accuracy (GAMA)}} & \textbf{Robust Accuracy (AutoAttack)} \\ 
\midrule
\multicolumn{1}{c|}{TRADES-AWP}                  &          & 62.73               & 29.92                       & 29.59                                 \\ 
\multicolumn{1}{c|}{DAJAT (\textbf{Ours})}                                      & WRN-34-10                                                      & 68.74                                       & 31.58                                               & 31.30                                 \\
\multicolumn{1}{c|}{\textbf{Gains using DAJAT}}            &  & \textbf{6.01}                               & \textbf{1.66}                                       & \textbf{1.71}                         \\
\midrule
\multicolumn{1}{c|}{TRADES-AWP}                                      &                                                      & 63.12                                       & 30.15                                               & 29.83                                 \\
\multicolumn{1}{c|}{DAJAT (\textbf{Ours})}                                   & WRN-34-20                                                      & 70.49                                       & 32.91                                               & 32.55                                 \\
\multicolumn{1}{c|}{\textbf{Gains using DAJAT}}                        &                         & \textbf{7.37}                               & \textbf{2.76}                                       & \textbf{2.72}      \\
\bottomrule
\end{tabular}}
\vspace{-0.1cm}
\end{table}
\begin{table}[t]
\centering
\caption{Comparison of DAJAT with Fixed-$\varepsilon$ and Ascending-$\varepsilon$ schedules on ResNet-18 and WideResNet-34-10 architectures on CIFAR-10 and CIFAR-100 datasets.}
\vspace{0.1cm}
\setlength\tabcolsep{2pt}
\resizebox{1.0\linewidth}{!}{
\label{table:vareps_abl}
\begin{tabular}{l|c|c|c|c|c|c}
\toprule
\multicolumn{1}{c|}{\textbf{\begin{tabular}[c]{@{}c@{}}Model\\ architecture\end{tabular}}} & \multicolumn{1}{c|}{\textbf{Dataset}}                 & \textbf{\begin{tabular}[c]{@{}c@{}}Epsilon \\ schedule\end{tabular}} & \textbf{Clean Acc} & \textbf{Robust Acc} & \textbf{\begin{tabular}[c]{@{}c@{}}Clean Acc \\ (with Weight Averaging)\end{tabular}} & \textbf{\begin{tabular}[c]{@{}c@{}}Robust Acc \\ (with Weight Averaging)\end{tabular}} \\ 
\midrule
                               & \multicolumn{1}{c|}{CIFAR-10} & Fixed $\varepsilon$                                                  & 86.57                                      & 51.17                                       & 86.25                                                                                                         & 51.44                                                                                                          \\ 

                                                                                  & CIFAR-10                                              & Varying $\varepsilon$ (DAJAT)                                        & 86.13                                      & 51.37                                       & 85.99                                                                                                         & 51.71                                                                                                          \\

\textbf{ResNet-18}                                                                          & CIFAR-100                                    & Fixed $\varepsilon$                                         & {66.03}                             & {26.24}                              & 65.50                                                                                                         & 26.55                                                                                                          \\

                                                                                  & CIFAR-100                                             & Varying $\varepsilon$ (DAJAT)                                        & 66.50                                      & 27.12                                       & 66.84                                                                                                         & 27.61                                                                                                          \\
\midrule

                                                                                   & CIFAR-10                                              & Fixed $\varepsilon$                                                  & 91.46                                      & 31.01                                       & 90.09                                                                                                         & 47.30                                                                                                          \\

                                                                          & CIFAR-10                               &Varying $\varepsilon$ (DAJAT)                               & {89.12}                             & {56.98}                              & 88.90                                                                                                         & 57.22                                                                                                          \\

\textbf{WRN-34-10}                                                                                 & CIFAR-100                                             & Fixed $\varepsilon$                                                  & 71.04                                      & 19.90                                       & 70.60                                                                                                         & 25.97                                                                                                          \\
                                                                                  & CIFAR-100                                             & Varying $\varepsilon$ (DAJAT)                                        & 68.82                                      & 30.75                                       & 68.74                                                                                                         & 31.58      \\
\bottomrule
\end{tabular}}
\vspace{-0.1cm}
\end{table}

\begin{figure}

\centering
        \includegraphics[width=\linewidth]{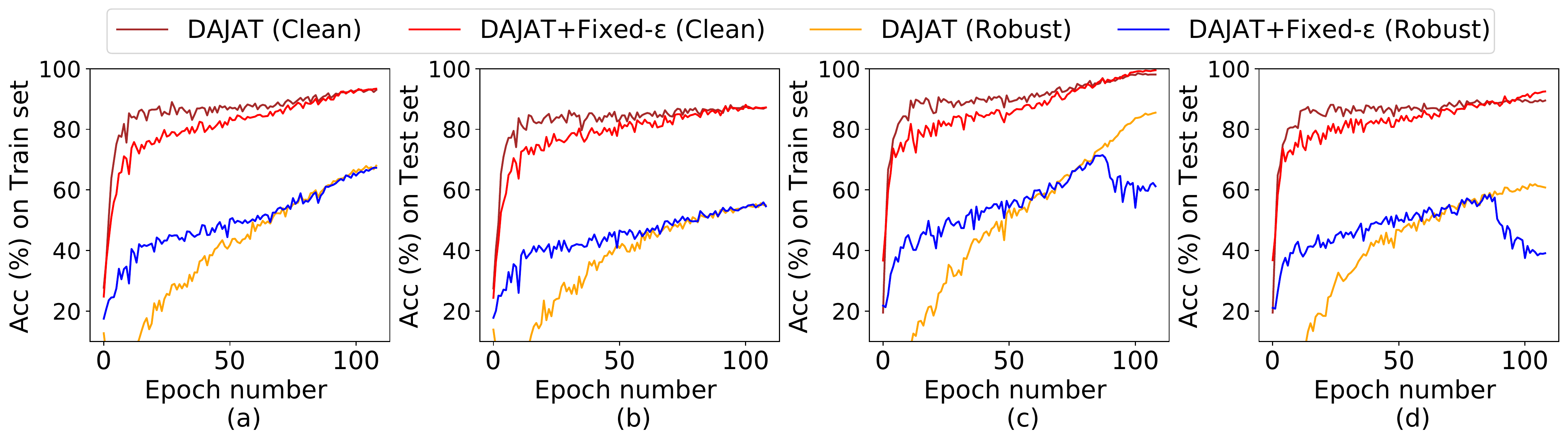}
        \caption{\textbf{Train and test Accuracy plots of DAJAT and DAJAT+Fixed-$\varepsilon$ to show impact of varying $\varepsilon$ training schedule:} The train and test plots of DAJAT are compared with an ablation experiment where a fixed $\varepsilon$ training schedule is used (DAJAT+Fixed-$\varepsilon$). Clean and Robust accuracy against PGD-20 \cite{madry-iclr-2018} attack are plotted on CIFAR-10 dataset using ResNet-18 (a,b) and WideResNet-34-10 (c,d) architectures. Although DAJAT+Fixed-$\varepsilon$ is stable at lower model capacities (ResNet-18), there is a sudden drop in the robust accuracy due to Gradient masking on WideResNet-34-10 (c,d) after epoch 87, indicating the need for using varying $\varepsilon$ schedule in DAJAT. Evaluations are done using an attack perturbation bound of $\varepsilon=8/255$.}
        \label{fig:co}
\end{figure}

We find that the performance gains obtained using DAJAT are indeed higher on larger capacity models. In Table-\ref{table:SOTA_WRN3420}, we present results on CIFAR-100 dataset, on WideResNet-34-10 and WideResNet-34-20 model architectures using 110 epochs of training. We note that the improvements in clean accuracy increase from 6.01\% to 7.37\%, while the improvements in robust accuracy against AutoAttack increase from 1.71\% to 2.72\%.
\subsection{Impact of Ascending Perturbation Radius in DAJAT}
\label{app:dajat-fixed}
We present an ablation of DAJAT without ascending perturbation radius (DAJAT+Fixed-$\varepsilon$), where attacks are constrained within a fixed perturbation bound of 8/255 during training. Table-\ref{table:vareps_abl} shows results on CIFAR-10 and CIFAR-100 datasets with WideResNet-34-10 and ResNet-18 architectures. Robust Accuracy is reported against the GAMA attack. As seen in the first half of the table, on ResNet-18 architecture, there is no gradient masking even with DAJAT+Fixed-$\varepsilon$. However, on the WideResNet-34-10 architecture, DAJAT+Fixed-$\varepsilon$ results in a large drop in robust accuracy due to gradient masking. Although the use of weight averaging improves results, its robust accuracy is still lower by around 5-10\% when compared to DAJAT. The phenomenon of gradient masking in DAJAT+Fixed-$\varepsilon$ can also be observed in Fig.\ref{fig:co}, where its robust accuracy suddenly drops after epoch 87 accompanied by an increase in clean accuracy on CIFAR-10 with WideResNet-34-10 architecture. We note that even with respect to the best epoch accuracy on DAJAT+Fixed-$\varepsilon$, we obtain an improvement of 2.57\% on robust accuracy using DAJAT.

\subsection{Variance across reruns}
\begin{table}[t]
\centering
\caption{\textbf{Variance across reruns:} Variation in performance (\%) of the proposed defenses ACAT and DAJAT on CIFAR-10 dataset and ResNet-18 architecture across three reruns. Based on the low standard deviation across runs, we note that both approaches are stable across reruns. Robust accuracy is evaluated against GAMA PGD-100 attack \cite{sriramanan2020gama}.}
\vspace{0.1cm}
\setlength\tabcolsep{2pt}
\label{table:variance}
\begin{tabular}{l|c|c||c|c}
\toprule
\multirow{2}{*}{}         & \multicolumn{2}{c||}{\textbf{ACAT}}                                       & \multicolumn{2}{c}{\textbf{DAJAT}}                                      \\
\cmidrule{2-5}
                          & \multicolumn{1}{l|}{\textbf{Clean Acc}} & \multicolumn{1}{l||}{\textbf{Robust Acc}} & \multicolumn{1}{l|}{\textbf{Clean Acc}} & \multicolumn{1}{l}{\textbf{Robust Acc}} \\
                          \midrule
Run-1 & 82.41                         & 50.00                          & 86.13                         & 51.37                          \\
Run-2 & 82.49                         & 50.08                          & 85.96                         & 51.51                          \\
Run-3 & 82.54                         & 50.12                          & 85.94                         & 51.48                          \\
\midrule
Average                   & 82.48                         & 50.07                          & 86.01                         & 51.45                          \\
Standard Deviation                  & 0.07                          & 0.06                           & 0.1                           & 0.07     \\
\bottomrule

\end{tabular}
\end{table}
\begin{table}[t]
\centering
\caption{\textbf{Comparison of compute, FLOPs (per iteration) and performance} of the proposed approaches DAJAT and ACAT when compared to TRADES-AWP on CIFAR-10 dataset.}
\vspace{0.1cm}
\setlength\tabcolsep{2pt}
\resizebox{1.0\linewidth}{!}{
\label{table:flops_WRN}
\begin{tabular}{l|c|c|c|c|c|c|c|c}
\toprule
\multicolumn{1}{c|}{\textbf{Model}} &
\multicolumn{1}{c|}{\textbf{Method}}                    & \multicolumn{2}{c|}{\textbf{gigaFLOPS (GFLOPS)}}        &
\multicolumn{2}{c|}{\textbf{Parameters (Million)}}  &\multicolumn{1}{c|}{\textbf{Train Time}} & \multicolumn{1}{c|}{\textbf{Clean Acc}} & \multicolumn{1}{c}{\textbf{Robust Acc}}  \\
\cline{3-6}
\rule{0pt}{2.5ex}& & {Training} & {Inference} & {Training} & {Inference} & \textbf{/epoch(sec)} & \textbf{(\%)} & \textbf{(\%)} \\
\midrule &
\multicolumn{1}{c|}{TRADES-AWP} &2707.6   & {71.251}                & 11.174                                 & 11.174                                  & 299                                & 80.47                  & 49.87                   \\ 
 ResNet18 &
ACAT (\textbf{Ours})                                                   & 997.52                       & 	   71.251                     &    11.174                                 & 11.174                                  & 108                                & 82.41                  & 49.80                    \\ &
DAJAT (\textbf{Ours})                                                     & 2422.5                                          &    71.251                 & 11.184                                 & 11.174                                  & 264                                & 85.99                  & 51.48          \\

\midrule &
\multicolumn{1}{c|}{TRADES-AWP} & 32417  &        853.08         & 46.160                                 & 46.160                                  & 1633                               & 85.10                  & 55.87                   \\ 
WRN-34-10~~~  &
ACAT (\textbf{Ours})                                                   & 11943
                                                                  &  853.08               & 46.160                                 & 46.160                                  & 472                                & 86.71                  & 55.36                   \\
&
DAJAT (\textbf{Ours})                                                     &      29005                                            & 853.08                & 46.183                              & 46.160                                  & 1381                               & 88.90                  & 56.96        \\
\bottomrule
\end{tabular}}
\end{table}
We present the variation across three reruns for the proposed defenses ACAT and DAJAT (Base, 2*AA) on CIFAR-10 dataset and ResNet-18 architecture in Table-\ref{table:variance}. Since weight averaging is known to improve the stability of the base method \cite{sriramanan2021nuat}, we present results without the use of weight averaging to highlight the inherent variance of the base algorithm. The standard deviation of robust accuracy across reruns is low (0.06-0.07) across both approaches indicating stability of the proposed method. It can be noted that the standard deviation of clean accuracy in the proposed defense DAJAT is slightly higher (0.1) due to the randomness in the complex augmentations, which impacts clean accuracy more than the robust accuracy. Overall, we note that the standard deviation of clean and robust accuracy for both proposed defences ACAT and DAJAT is low . 

\subsection{Analysis on Compute, Flops and Performance}
\label{app:flops}
We present the FLOPs, number of parameters and training time on a single Nvidia RTX-3090 GPU along with our results when compared to the TRADES-AWP baseline on the CIFAR-10 dataset for ResNet18 and WRN-34-10 models in Table-\ref{table:flops_WRN}. Robust accuracy is reported against AutoAttack. We discuss our observations below:
\begin{itemize}
\item  FLOPs and number of parameters during inference are identical among the three training methods (TRADES-AWP, ACAT, DAJAT), since we use only a single batch-normalization layer (corresponding to Pad+Crop) during inference. As expected, these values are higher for WideResNet-34-10 model architecture when compared to ResNet-18.
\item  Since we use split batch-norm layers in DAJAT, the number of parameters increases by 0.05\% during training, while it remains the same during inference.
\item  We compute FLOPs during training by considering that a single backward pass requires twice the number of FLOPs when compared to a forward pass. We also provide the number of forward and backward passes in each method for reference.
\item  Using ACAT, we achieve 63\% reduction in FLOPs (training) and training time when compared to the TRADES-AWP baseline, while achieving 1.6-1.9\% higher clean accuracy and comparable robust accuracy.
\item The use of ACAT strategy in the proposed DAJAT defense enables us to achieve similar computational complexity as TRADES-AWP defense, while obtaining gains in performance. Using DAJAT, we achieve 10\% reduction in FLOPs (training) and training time, while obtaining 3.8-5.5\% higher clean accuracy and 1-1.6\% higher robust accuracy.
\end{itemize}
\subsection{Comparison against CutMix based augmentation}
\begin{table}[t]
\centering
\caption{\textbf{Comparison of the proposed augmentation scheme with CutMix based augmentations \cite{rebuffi2021data}:} Performance (\%) of the proposed defense DAJAT (Base, 2*AA) when compared to the use of CutMix based augmentation proposed by Rebuffi et al. \cite{rebuffi2021data} against PGD 40-step attack \cite{madry-iclr-2018} }
\vspace{0.1cm}
\setlength\tabcolsep{2pt}
\label{table:deepmind}
\begin{tabular}{l|c|c}
\toprule

         \textbf{Method}                                 & \textbf{Clean Acc}                & \textbf{Robust Acc (PGD-40)}      \\
                                          \midrule
TRADES \cite{zhang2019theoretically}                           & 84.72                         & 56.92                \\
Rebuffi et al. \cite{rebuffi2021data}                   & 87.24                         & 57.60                \\
TRADES-AWP \cite{wu2020adversarial}                     & 85.35                         & 59.13                \\
DAJAT (\textbf{Ours}) (Base, 2*AA)                   & \textbf{88.90} & \textbf{60.97} \\
\bottomrule

\end{tabular}
\end{table}
\label{sec:cutmix_deepmind}
While we compare the performance of the proposed approach against various base adversarial training algorithms \cite{wu2020adversarial,madry-iclr-2018,pang2020bag,sriramanan2021nuat} in the main paper, we additionally compare with a recent augmentation scheme that uses CutMix augmentations \cite{rebuffi2021data} to improve performance in this section. The authors of \cite{rebuffi2021data} show a significant boost in performance using 400 epochs of training and large model architectures. However, to ensure a fair comparison, we report the result of 110 epochs of training on WideResNet-34-10 architecture and CIFAR-10 dataset that has been shared by the authors with us upon request. We report the PGD 40-step accuracy as shared by the authors. As shown in Table-\ref{table:deepmind}, we obtain a significant boost in performance over the CutMix based augmentation as well as the TRADES-AWP \cite{wu2020adversarial} baseline using the proposed defense DAJAT. 

Additionally, contrary to the claims by Rebuffi et al. \cite{rebuffi2021data}, we show that it is indeed possible to effectively use augmentations that modify the low-level statistics of images for obtaining improved performance in Adversarial Training by using the proposed defense DAJAT. 
\begin{table}
\centering
\caption{Performance (\%) by using \cite{rade2021pytorch} on CIFAR-10 dataset with Preact-ResNet18 model with Swish Activation trained using varying epsilon schedule and cosine learning rate unless specified otherwise.}
\vspace{0.1cm}
\label{table:cutmix}
\begin{tabular}{l|c|c}
\toprule
\multicolumn{1}{l|}{\textbf{Method}}               & \textbf{Clean}                & \textbf{GAMA}                 \\
 \midrule
{[}C1{]}: Cutmix - step schedule + fixed eps & 81.67                & 49.18                \\
{[}C2{]}: Cutmix                                                              & \textbf{83.34} & 49.24 \\
{[}C3{]}: Ours(Base, Cutmix)                                                  & 82.67                         & 51.99                         \\
{[}C4{]}: Ours(Base, Cutmix, Cutmix)                  & 83.05 &	\textbf{52.22}            \\
{[}C5{]}: C1 with Relu                                                      & 81.03                         & 46.6                          \\
{[}C6{]}: C1 with Weight decay for BN                                    & 70.66                         & 36.36 \\

\bottomrule

\end{tabular}
\end{table}

As present in the github repository \footnote{https://github.com/deepmind/deepmind-research/tree/master/adversarial\_robustness/pytorch} of \cite{rebuffi2021data} we note that naively using cutmix doesn't give good results as shown in Table-\ref{table:aug}, therefore as suggested we use \cite{rade2021pytorch} as the base code and incorporate cutmix into it. We present the results for 200 epochs training with learning rate drop of 0.1 at 100 and 150 epochs, using the PreActResNet-18 model with Swish Activation and batch size of 128 in Table-\ref{table:cutmix}(C1). We observe significantly improved results as compared to Table-\ref{table:aug} on using \cite{rade2021pytorch} as the base code. We observe that the key differences in \cite{rade2021pytorch} as compared to \cite{zhang2019theoretically} are:
\begin{itemize}
    \item Use of swish activation function in the PreActResNet18 model 
    \item Weight decay not used for batch normalization layers
\end{itemize}
To study the impact of these changes, we investigate the use ReLU instead of Swish activation (Table-\ref{table:cutmix}(C5)) and the use of weight decay for all the parameters of the model including the batch normalization layers (Table-\ref{table:cutmix}(C6)). In both cases, we observe a significant drop with respect to C1. Thus based on this ablation, the use of swish activation, and avoiding weight decay for batch normalization layers seems important to obtain benefits using Cutmix.

Further we incorporate linearly increasing varying epsilon schedule along with cosine learning rate schedule and get improved results in Table-\ref{table:cutmix}(C2). Next we incorporated our method DAJAT with C2 and present the results in Table-\ref{table:cutmix}(C3,C5), where we can observe significant gains over C2, thus showing the effectiveness of DAJAT.

\begin{table*}[t]
\caption{Performance of the proposed defense DAJAT when compared to some concurrent works on \textbf{CIFAR-10 and CIFAR-100 datasets} for ResNet18 and WideResNet-34-10 models. Robust evaluations are performed on Auto-Attack \cite{croce2020reliable}.}

\setlength\tabcolsep{2pt}
\resizebox{1.0\linewidth}{!}{
\label{table:contemporary_works}
\begin{tabular}{l|c|c||c|c||c|c||c|c}
\toprule
\multicolumn{1}{l|}{\textbf{}} & \multicolumn{2}{c||}{\textbf{CIFAR-10, ResNet-18}} & \multicolumn{2}{c||}{\textbf{CIFAR-10, WRN-34-10}} & \multicolumn{2}{c||}{\textbf{CIFAR-100, ResNet-18}} & \multicolumn{2}{c}{\textbf{CIFAR-100, WRN-34-10}} \\
 \cmidrule{2-9}
\multicolumn{1}{l|}{\textbf{Method}}          & \textbf{Clean}                      & \textbf{AA @ 8/255}                & \textbf{Clean}                      & \textbf{AA @ 8/255}                & \textbf{Clean}                    & \textbf{AA @ 8/255}                   & \textbf{Clean}               & \textbf{AA @ 8/255}                        \\
\midrule

AWP \cite{wu2020adversarial}                                                   & 81.99                               & 51.45                              & 85.36                               & 56.17                              & 59.88                             & 25.81                                 & 62.73                        & 29.59                                      \\
HAT \cite{rade2022reducing}                                                  & 85.63                               & 49.54                              & 86.21                               & 51.46                              & 59.19                             & 23.26                                 & 59.95                        & 24.55                                      \\
SEAT \cite{wang2022selfensemble}                                                  & 83.7                                & 51.3                               & 86.44                               & 55.67                              & 56.28                             & \textbf{27.87}                                 & -                           & -                                         \\
SEAT+Cutmix \cite{wang2022selfensemble}                                           & 81.53                               & 49.1                               & 84.81                               & 56.03                              & -                                & -                                    & -                           & -                                         \\
TRADES + TE \cite{dong2022exploring}                                         & 83.86                               & 49.77                              & -                                  & -                                 & 59.35                             & 25.27                                 & -                           & -                                         \\
UDR + TRADES \cite{bui2022a}                                         & 84.4                                & 49.9                               & 84.93                               & 54.45                              & -                                & -                                    & -                           & -                                         \\

DAJAT (\textbf{Ours})                                                & \textbf{85.71}       & \textbf{52.50}      & \textbf{88.71}       & \textbf{57.81}      & \textbf{65.45}                             & 27.69                                 & \textbf{68.75}                        & \textbf{31.85}              \\

\bottomrule
\end{tabular}}
\vspace{-0.1cm}
\end{table*}

\vspace{-0.2cm}

\subsection{Comparison of DAJAT with Concurrent Works}
Here we compare the proposed approach DAJAT trained for 200 epochs with recent works that appeared at ICLR 2022.
The comparison of the proposed method DAJAT with AWP \cite{wu2020adversarial}, HAT \cite{rade2022reducing}, Self Ensemble Adversarial Training (SEAT) \cite{wang2022selfensemble} with (SEAT+cutmix) and without cutmix (SEAT), Unified distributional robustness for TRADES (UDR + TRADES) \cite{bui2022a}, temporal ensembling with TRADES (TRADES + TE) \cite{dong2022exploring} on CIFAR10 and CIFAR100 datasets for ResNet18 and WideResNet-34-10 models is shown in Table-\ref{table:contemporary_works}. Since these are very recent works, we only present results reported in the paper, and leave the remaining entries in the table blank. 
\begin{table*}[t]
\caption{Performance (\%) of DAJAT when \textbf{combined with other Adversarial training methods}, OAAT \cite{addepalli2021oaat} and HAT \cite{rade2022reducing} on \textbf{CIFAR-10 and CIFAR-100} with 110 epochs of training. Robust evaluations are performed on Auto-Attack(AA) \cite{croce2020reliable} at $\varepsilon=8/255$ and $16/255$.}
\vspace{0.1cm}
\setlength\tabcolsep{2pt}
\resizebox{1.0\linewidth}{!}{
\label{table:dajat_with_others}
\begin{tabular}{l|c|c|c||c|c|c||c|c|c||c|c|c}
\toprule
\multicolumn{1}{l|}{} & \multicolumn{3}{c||}{\textbf{CIFAR-10, ResNet-18}} & \multicolumn{3}{c||}{\textbf{CIFAR-10, WRN-34-10}}                                                                           & \multicolumn{3}{c||}{\textbf{CIFAR-100, ResNet-18}}                   & \multicolumn{3}{c}{\textbf{CIFAR-100, WRN-34-10}}                                                                          \\
\cmidrule{2-13}
\multicolumn{1}{l|}{\textbf{Method}} & \textbf{Clean}      & \textbf{AA, 8/255}     & \textbf{AA, 16/255}     & \textbf{Clean}                           & \textbf{AA, 8/255}                          & \textbf{AA, 16/255}                                     & \textbf{Clean}                    & \textbf{AA, 8/255}                   & \textbf{AA, 16/255}             & \textbf{Clean}                           & \textbf{AA, 8/255}                          & \textbf{AA, 16/255}                                     \\
\midrule
AWP \cite{zhang2019theoretically,wu2020adversarial} & 80.47 & 49.87 & \textbf{19.23} & 85.10                         & 55.87                         & \textbf{23.27}                         & 59.88 & 25.81 & 8.28  & 62.73                         & 29.59                         & \textbf{11.04}                        \\
AWP+DAJAT    & \textbf{85.99}     & \textbf{51.48} & 16.33 & \textbf{88.90}                         & \textbf{56.96}                         & 19.73                         & \textbf{66.84} & \textbf{27.32} & \textbf{8.97}  & \textbf{68.74}                         & \textbf{31.30}                         & 9.91                         \\
\midrule
OAAT  \cite{addepalli2021oaat}                               & 80.24                                  & 50.88 & 22.05 & 85.67                         & 55.93                         & 24.05                         & 61.70 & 26.77 & 9.91  & 65.73                         & 30.35                         & 12.01                        \\
OAAT+DAJAT                           & \textbf{82.05}                                  & \textbf{52.21} & \textbf{22.78} & \textbf{86.22}                         & \textbf{57.64}                         & \textbf{24.56}                         & \textbf{62.50} & \textbf{28.47} & \textbf{10.67} & \textbf{66.03}                         & \textbf{31.15}                         & \textbf{12.67}                        \\
\midrule
HAT   \cite{rade2022reducing}                               & 85.63                                  & 49.54 & 14.96 & 86.21                         & 51.46                         & \textbf{16.76}                         & 59.19 & 23.26 & 6.96  & 59.95                         & 24.55                         & 7.13                         \\
HAT+DAJAT                            & \textbf{86.68}                                  & \textbf{51.47} & \textbf{16.38} & \textbf{86.71} & \textbf{53.85} & 16.50 & \textbf{62.78} & \textbf{26.49} & \textbf{8.72}  & \textbf{64.88} & \textbf{27.37} & \textbf{8.71}
\\
\bottomrule
\end{tabular}}
\end{table*}

\begin{figure}
\centering
\includegraphics[width=\linewidth]{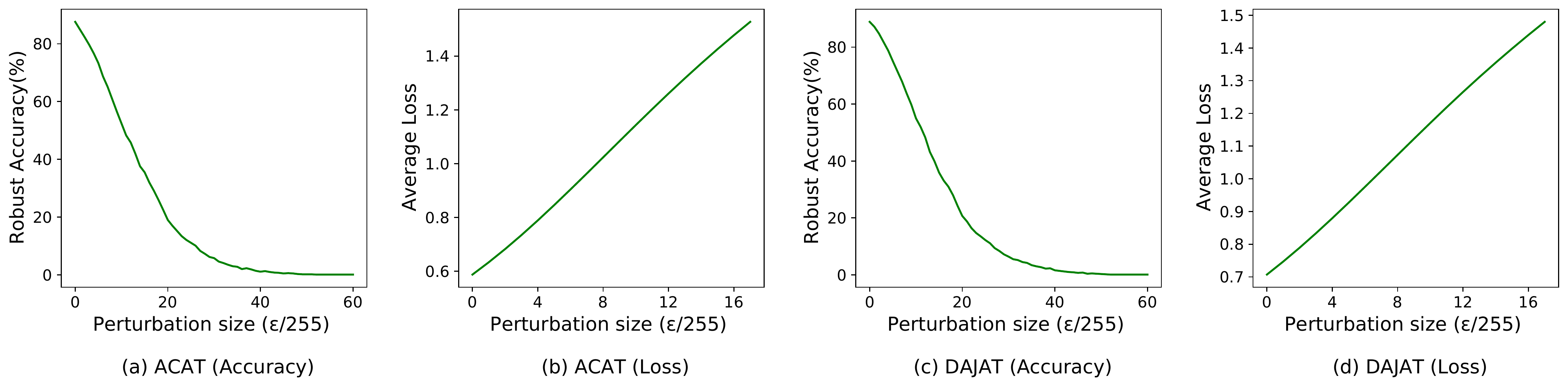}
\vspace{-0.4cm}
\caption{\textbf{Robust Accuracy and Loss on FGSM attack against variation in perturbation size:} (a,c) Robust accuracy (\%) of the proposed defenses ACAT and DAJAT against PGD 7-step attacks across variation in attack perturbation bound. Attacks within larger perturbation bounds are able to bring down the robust accuracy of the model to 0, indicating the absence of gradient masking. (b,d) Cross-entropy loss on FGSM adversarial samples across variation in attack perturbation bound. The linearly increasing trend of loss indicates the absence of gradient masking. The models are trained on CIFAR-10 dataset using ResNet-18 architecture.}
\vspace{-0.4cm}
\label{fig:loss_plot}
\end{figure}
\begin{table}[t]
\centering
\caption{\textbf{Evaluation against Black-Box and White-Box FGSM \cite{goodfellow2014explaining} attacks and multi-step PGD attacks \cite{madry-iclr-2018}:} Performance (\%) of the proposed DAJAT (Base + 2*AA) defense on CIFAR-10 dataset with ResNet-18 architecture}
\vspace{0.1cm}
\setlength\tabcolsep{2pt}
\label{table:pgd_fgsm}
\begin{tabular}{l|c|c|c|c|c|c}
\toprule
\textbf{Method} & \textbf{Clean Acc} & \textbf{BB FGSM} & \textbf{FGSM} & \textbf{PGD-20} & \textbf{PGD-100} & \textbf{PGD-500} \\
\midrule
NuAT2-WA             & 86.32              & 84.71            & 63.48         & 58.09           & 57.74            & 57.74            \\
ACAT                 & 86.71              & 85.29            & 64.08         & 58.76           & 58.64            & 58.53            \\
TRADES-AWP                  & 85.36              & 83.93            & 63.49         & 59.22           & 59.11            & 59.08            \\
DAJAT(Base, 3*AA )   & \textbf{88.64}              & \textbf{87.19}            & \textbf{66.99}         & \textbf{61.09}           & \textbf{60.80}             & \textbf{60.74}  \\
\bottomrule
\end{tabular}
\vspace{-0.1cm}
\end{table}

\begin{table}[t]
\centering
\caption{\textbf{Evaluation against multi-step Targeted and Untargeted PGD attacks \cite{madry-iclr-2018} with single and multiple random restarts:} Performance (\%) of the proposed defense DAJAT (Base, 2*AA) across different datasets with ResNet-18 architecture}
\vspace{0.1cm}
\setlength\tabcolsep{2pt}
\label{table:rand_restarts}
\begin{tabular}{l|c|c|c|c|c|c}
\toprule
 \multicolumn{1}{l|}{}        & \multicolumn{2}{c|}{\textbf{CIFAR-10}} & \multicolumn{2}{c|}{\textbf{CIFAR-100}} & \multicolumn{2}{c}{\textbf{IN-10}} \\
\cmidrule{2-7}
\textbf{Attack}   & 500-step     & 1000-step     & 500-step      & 1000-step     & 500-step    & 1000-step   \\
                                  \midrule
PGD-Targeted (Least Likely Class) & 85.01        & 85.01         & 66.02         & 65.98         & 85.06       & 85.01       \\
PGD-Targeted (Random Class)       & 80.56        & 80.55         & 63.96         & 63.96         & 80.13       & 80.13       \\
PGD-Untargeted                    & 55.21        & 55.20         & 32.89         & 32.89         & 65.07       & 65.07       \\
\midrule
\multicolumn{1}{l|}{}              & 1-RR         & 1000-RR       & 1-RR          & 1000-RR       & 1-RR        & 1000-RR     \\
\midrule
PGD 50-step, r-RR                 & 55.30        & 54.55         & 32.98         & 32.09         & 65.20       & 65.02  \\    
\bottomrule
\end{tabular}
\end{table}

\begin{figure}
\centering
\includegraphics[width=\linewidth]{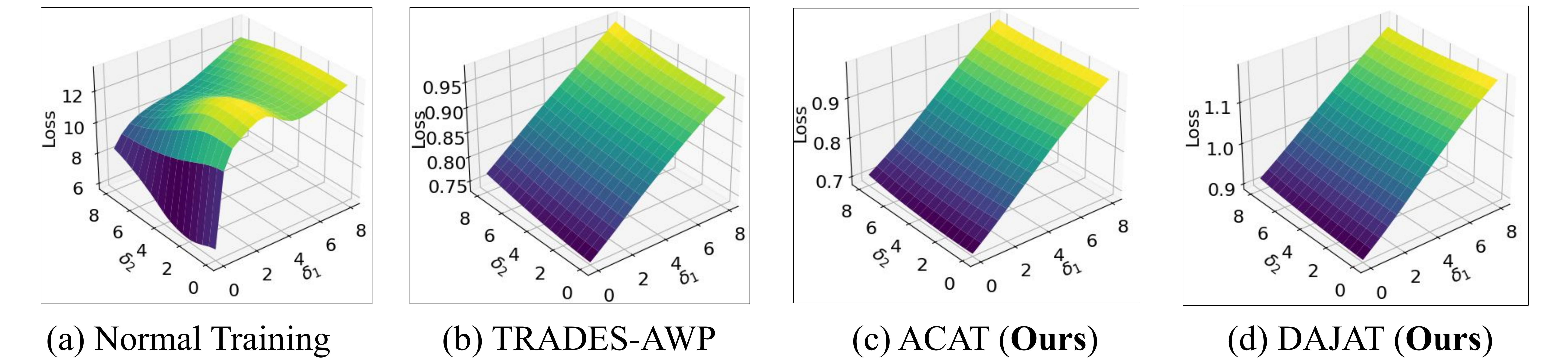}
\vspace{-0.4cm}
\caption{\textbf{Loss Surface Plots:} Plot of cross-entropy loss in the local neighborhood of images along the gradient direction ($\delta_1$) and a random direction perpendicular to the gradient ($\delta_2$). The loss surface of the proposed defenses ACAT and DAJAT are smooth similar to the TRADES-AWP defense, indicating the absence of gradient masking.}
\vspace{-0.5cm}
\label{fig:loss_surface_plot}
\end{figure}

\subsection{Combining the proposed approach with different adversarial training methods}
We explore combining the proposed defense DAJAT with some existing methods like \cite{wu2020adversarial}, \cite{addepalli2021oaat} and \cite{rade2022reducing} in Table-\ref{table:dajat_with_others}. We observe that combining DAJAT with all three existing works leads to significant gains both in clean as well as adversarial accuracies (AA, 8/255), especially on CIFAR-100 where the number of images per class is low. Although OAAT \cite{addepalli2021oaat} shows improved results over AWP \cite{wu2020adversarial}, combining DAJAT with OAAT leads to further gains of $\sim1.5\%$ in both clean and adversarial accuracy on CIFAR10 and  $1-1.5\%$ gains in both clean and adversarial accuracy on CIFAR100. Further, since OAAT \cite{addepalli2021oaat} claims to achieve robustness at larger epsilon bounds, we evaluate using Auto-Attack at $\varepsilon=16/255$. Using OAAT+DAJAT we observe gains over OAAT on AutoAttack with $\varepsilon=16/255$ as well, which further confirms the effectiveness of DAJAT. Finally we combine DAJAT with HAT \cite{rade2022reducing} and we observe consistent gains over HAT \cite{rade2022reducing} on all models and datasets. While HAT \cite{rade2022reducing} proposes to improve the robustness-accuracy trade-off, combining DAJAT with HAT further improves this trade-off and shows gains of $\sim1\%$ on clean accuracy and $\sim2\%$ on robust accuracy for CIFAR-10, and $3-5\%$ on clean accuracy and $\sim3\%$ on robust accuracy for CIFAR-100 dataset. 
\subsection{Sanity checks to verify the absence of gradient masking}    

We perform several sanity checks as recommended by Athalye et al. \cite{athalye2018obfuscated} to ensure the absence of gradient masking in the proposed defenses ACAT and DAJAT. 

\begin{itemize}
\itemsep0em
\vspace{-0.1cm}
\item From Table-\ref{table:pgd_fgsm} we note that Black-Box attacks are weaker than White-Box attacks, indicating that the gradients from the model are reliable.
\item We further note from Table-\ref{table:pgd_fgsm} that attacks with higher number of steps are stronger than those with lower steps. Further, PGD multi-step attacks are stronger than FGSM white-box attacks. 
\item From Table-\ref{table:rand_restarts} we note that robust accuracy against targeted and untargeted attacks saturates as the number of attack steps increase from 500 to 1000, indicating that the evaluation is robust.
\item We also note from Table-\ref{table:rand_restarts} that the drop in accuracy with 1000 random restarts is marginal. 
\item We note from Fig.\ref{fig:loss_plot} that an increase in perturbation bound increases the effectiveness of PGD 7-step attacks, and is able to bring down the accuracy of the model to 0 at large bounds. Further, the loss on FGSM samples monotonically increases in the vicinity of the data samples. These trends indicate the absence of gradient masking. 
\item We present results against AutoAttack \cite{croce2020reliable} in Tables-\ref{table:sota_c10} and \ref{table:sota_c100} of the main paper. AutoAttack is an ensemble of several gradient-based attacks and a gradient-free attack Square \cite{andriushchenko2019square}. The robust accuracy against AutoAttack is similar to the accuracy against gradient-based attack GAMA \cite{sriramanan2020gama} indicating that gradient-free attacks are not significantly stronger than gradient based attacks. 
\item We show the loss surface plots of the proposed defenses ACAT and DAJAT in the vicinity of data samples in Fig.\ref{fig:loss_surface_plot}. We note that the loss surface of the proposed defenses is smooth similar to the TRADES-AWP defense, indicating the absence of gradient masking.
\end{itemize}

We finally compare the robust accuracy against various attacks in Tables-\ref{table:pgd_fgsm} and \ref{table:rand_restarts} with the robust accuracy against GAMA attack \cite{sriramanan2020gama} and AutoAttack \cite{croce2020reliable} in Tables-\ref{table:sota_c10} and \ref{table:sota_c100} in the main paper. The latter evaluations are significantly stronger, indicating that the evaluation presented in the main paper is robust. 
\end{document}